%% file: main.tex
\documentclass[twoside]{article}

\usepackage[accepted]{aistats2025}
\usepackage{booktabs}       
\usepackage{amsfonts}       
\usepackage{nicefrac}       
\usepackage{microtype}      
\usepackage{xcolor}         

\usepackage[ruled,vlined]{algorithm2e}
\usepackage{algorithmic}
\usepackage{amsmath, amsthm, amssymb,amsfonts}

\usepackage{graphicx,psfrag,epsf}
\usepackage{enumerate}
\usepackage{bm}
\usepackage[style=authoryear]{biblatex}
\usepackage{xr-hyper}
\usepackage{hyperref}  
\usepackage{cleveref}
\crefname{section}{Section}{Sections}
\crefname{table}{Table}{Tables}
\crefname{figure}{Figure}{Figures}
\crefname{appendix}{Appendix}{Appendices}
\crefname{prop}{Proposition}{Propositions}

\newtheorem{prop}{Proposition}

\usepackage{titletoc}
\titlecontents{section}
    [2em] 
    {\vspace{.5em}\bfseries} 
    {\contentslabel{1.5em}} 
    {} 
    {\titlerule*[.5em]{.}\contentspage} 

\begin{document}

%
\runningtitle{Deep Vecchia Ensemble}

%

\twocolumn[

\aistatstitle{Vecchia Gaussian Process Ensembles on Internal Representations of Deep Neural Networks}

\aistatsauthor{ Felix Jimenez \And  Matthias Katzfuss }

\aistatsaddress{University of Wisconsin--Madison \And University of Wisconsin--Madison } 
]

\input{contents/00_abstract}
\input{contents/01_intro}
\input{contents/02_background}
\input{contents/03_methods}
\input{contents/04_applications}
\input{contents/05_experiments}
\input{contents/06_conclusion}

\printbibliography
\newpage
\appendix
\clearpage
\setcounter{page}{1}
\onecolumn
\startcontents[appendix]
\aistatstitle{Supplementary Materials}
\printcontents[appendix]{l}{1}{\textbf{Supplementary Material Contents:}}
\input{contents/07_appendix}



\end{document}

%% file: contents/00_abstract.tex
\begin{abstract}
For regression tasks, standard Gaussian processes (GPs) provide natural uncertainty quantification (UQ), while deep neural networks (DNNs) excel at representation learning. Deterministic UQ methods for neural networks have successfully combined the two and require only a single pass through the neural network. However, current methods necessitate changes to network training to address feature collapse, where unique inputs map to identical feature vectors. We propose an alternative solution, the deep Vecchia ensemble (DVE), which allows deterministic UQ to work in the presence of feature collapse, negating the need for network retraining. DVE comprises an ensemble of GPs built on hidden-layer outputs of a DNN, achieving scalability via Vecchia approximations that leverage nearest-neighbor conditional independence. DVE is compatible with pretrained networks and incurs low computational overhead. We demonstrate DVE's utility on several datasets and carry out experiments to understand the inner workings of the proposed method.
\end{abstract}

%% file: contents/01_intro.tex
\section{INTRODUCTION}
In recent years, deep neural networks (DNNs) have achieved remarkable success in various tasks such as image recognition, natural language processing, and speech recognition. However, despite their excellent performance, these models have certain limitations, such as their lack of uncertainty quantification (UQ). Much of UQ for DNNs is based on a Bayesian approach that models network weights as random variables \parencite{neal2012bayesiandl} or involves ensembles of networks \parencite{lakshminarayanan2017deepensemble}, both requiring large memory and several evaluations of each test point. Deterministic single-forward-pass methods \parencite{amersfoot2020DUQ, liu2020sngp, mukhoti2023ddu} have emerged to avoid the need to evaluate multiple networks. Deterministic methods require changes to the network training procedure to generate well-structured feature spaces for use with an outside model, often a Gaussian process. 

Gaussian processes (GPs) provide natural UQ, but they are known to scale poorly with large datasets, which is addressed by myriad GP approximations. One such method is the Vecchia approximation \parencite{vecchia1988estimation, katzfuss2021general}, which uses nearest-neighbor conditioning sets to exploit conditional independence among the data. However, like all GP based methods, the Vecchia approximation requires that the input space is well structured. Therefore, naively applying a Vecchia GP with a deterministic method wouldn't address feature collapse.

\begin{figure}
    \centering
    \includegraphics[width =1.0\linewidth]{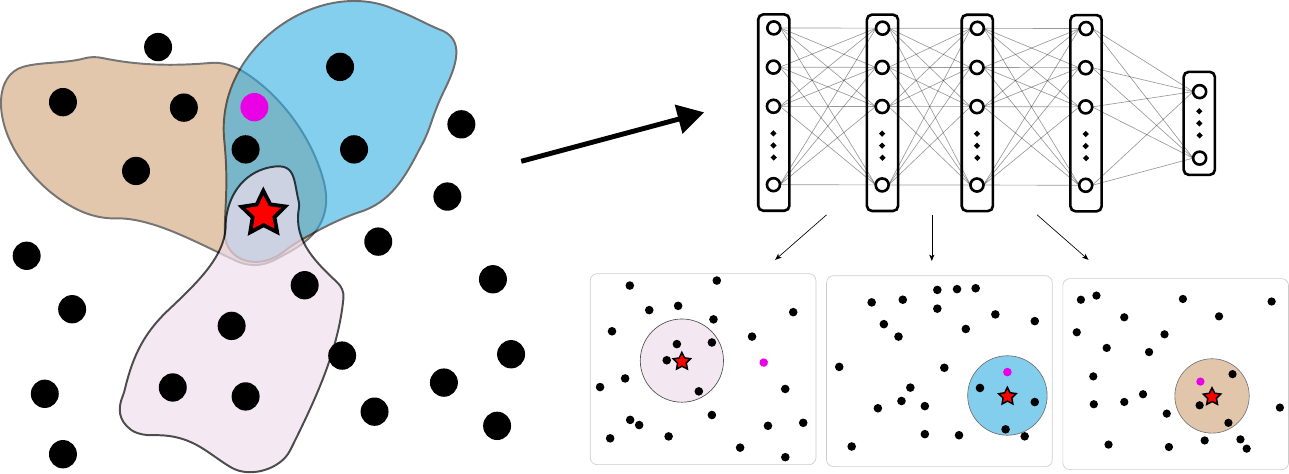}
    \caption{\textbf{Different layers imply different nearest neighbors}. Left: The input (red star) has different nearest-neighbor conditioning sets based on the metrics induced by the layers of the DNN with the magenta point being in two of the three conditioning sets. The brown, blue, and pink shaded areas denote the regions in input space that will be mapped to a hypersphere in the first, second, and third intermediate spaces, respectively. The conditioning sets derived from the different regions may overlap, as in the blue and brown region, or be disjoint from the others as in the pink region. Right: The labeled training data are propagated through the network and intermediate feature maps are stored. For a red test point, we assess uncertainty by considering and weighting instances in the training data that are similar to the test sequence in one or more of the feature maps.}
    \label{fig:intro_figure}
\end{figure}

The primary contribution of our paper is the introduction of the deep Vecchia ensemble (DVE), which leverages outputs from multiple internal DNN layers for deterministic uncertainty quantification without modifying network training. A high-level summary of our approach is given in \cref{fig:intro_figure}. Our method does not require changes to the standard DNN training regimen, unlike existing deterministic techniques. It offers interpretability by identifying training points similar to the query at test time, potentially enhancing performance in scenarios with limited data. Moreover, the DVE framework allows us to differentiate between aleatoric and epistemic uncertainty.

The remainder of the paper is organized as follows: We begin in \cref{sec:background} with a review of related concepts and define \textit{intermediate representations}. In \cref{sec:related_work}, we contextualize our approach within existing work on UQ for neural networks and GP approximation. The proposed methodology is detailed in \cref{sec:model_construction}. We then apply our model to pretrained neural networks in \cref{sec:applications}. This is followed by experiments in \cref{sec:experiments} that explore how our model works. In \cref{sec:conclusion}, we discuss potential application areas for our method, the limitations of our approach, and future research directions. 

%% file: contents/02_background.tex
\section{PRELIMINARIES}\label{sec:background}
\subsection{Modeling Functions with Gaussian Processes}\label{sec:background_gps}
Suppose we have a dataset $\mathcal{D} =\{\bm X, \bm y\} = \{\bm x_i, y_i\}_{i = 1}^n$, where $\bm x_i\in\mathcal{X}$, $y_i = f(\bm x_i) + \epsilon_i$, $\epsilon_i \stackrel{iid}{\sim} \mathcal{N}(0,\sigma^2)$, and $f:\mathcal{X}\rightarrow\mathbb{R}$. Our goal is to make predictions at $n_p$ test locations $\bm X^* = \{\bm x^*_j\}_{j = 1}^{n_p}$ in a probabilistic fashion. A common approach is to assign the latent function $f$ a zero-mean GP prior with covariance function $K_{\bm \theta}(\bm x, \bm x')$ parameterized by $\bm \theta$. Letting $\bm y^* = \{f(\bm x^*_j) + \epsilon^*_j\}_{j = 1}^{n_p}$, $p(\bm y^*, \bm y)$ will be jointly multivariate normal, which implies:
\begin{align}\label{eqn:f_test_posterior}
    p(\bm y^*|\bm y) = \mathcal{N}(\bm \mu^*, \bm K^*_{\bm \theta^*} + \sigma^2 \bm I_{n_p}),
\end{align}
where $\bm \mu^*$ and $\bm K^*_{\bm \theta^*}$ depend on $\mathcal{D}$ and $\bm \theta^*$ is selected to maximize the marginal log-likelihood:
\begin{align}\label{eqn:marginal_log_likelihood}
    p(\bm y) &= \int p(\bm y|\bm f)p(\bm f)d\bm f = \mathcal{N}(\bm 0, \bm K_{\bm \theta}(\bm X, \bm X) + \sigma^2 \bm I_n).
\end{align}

Both Equations \eqref{eqn:f_test_posterior} and \eqref{eqn:marginal_log_likelihood} have time and space complexities of $\mathcal{O}(n^3)$ and $\mathcal{O}(n^2)$, respectively.

\subsection{Approximating Gaussian Processes Using Vecchia}\label{sec:background_vecchia}
The joint distribution of a collection of random variables, $\bm y = \{y_1, y_2, ..., y_n\}$, can be decomposed as:

$$p(\bm y) = \prod_{i = 1}^n p(y_i|\bm y_{h(i)}), \hspace{.1cm} h(i) = \{1, ..., i-1\}$$ 

Vecchia's approximation \parencite{vecchia1988estimation} conditions the $i^{th}$ observation on a set indexed by $g(i)$, where $g(i) \subset h(i)$ with $|g(i)| \leq m$ for $m \ll n$. The approximate joint distribution is then, $p(\bm y) \approx \prod_{i = 1}^n p(y_i|\bm y_{g(i)})$. This approximation reduces the in-memory space complexity of Equation \eqref{eqn:marginal_log_likelihood} to $\mathcal{O}(n m)$. An additional approximation utilizing mini-batches \parencite{jimenez2022scalable, cao2022variableselection} of size $n_b$ results in an in-memory space complexity of $\mathcal{O}(n_b m^2)$ and a time complexity of $\mathcal{O}(n_b m^3)$.

For prediction, let $\tilde{\bm y} = (\bm y,\bm y^*)$, where $\bm y^* = \{y^*_{n+1}, ..., y^*_{n+n_p}\}$ are test points ordered after the observed data. The Vecchia approximation of the conditional distribution for $\bm y^*$ is, $p(\bm y^*|\bm y) \approx \prod_{j = n+1}^{n+n_p} p(y^*_j|\tilde{\bm y}_{g(j)})$, which reduces the time and space complexity of Equation \eqref{eqn:f_test_posterior} to $\mathcal{O}(n_p m)$. 

For the $i^{th}$ data point, the $m$ ordered nearest neighbors $g(i)$ are typically determined based on a distance metric, such as Euclidean distance, between input variables $\bm x_i$ and $\bm x_j$. To compute these ordered nearest neighbors, it is assumed that the data has been sorted according to some ordering procedure, and for each $i$, the $m$ nearest neighbors that precede $i$ in the ordering are selected.
\subsection{Ensembling Gaussian Processes}\label{sec:background_gp_ensembles}

Consider $L$ Gaussian processes meant to model the same function which we want to combine to form a single estimate. While we allow each GP to have its own hyperparameters, product-of-expert GP ensembles \parencite{tresp2000bcm, cao2014gpoe, deisenroth2015rbcm,rulliere2018nestedgp} typically share hyperparameters $\bm \theta$ between the GPs. We are assuming $\bm \theta_k$ contains all the hyperparameters in both the GP kernel and the likelihood. 

The prediction of $y^*$ at a location $\bm x^*$ can be formed by using a generalized product of experts (gPoE) \parencite{cao2014gpoe}, an extension of product of experts (PoE) \parencite{hinton2002poe}, which combines the predictions of $L$ models by using a weighted product of the predictive densities. Letting $p_k(y^*)$ denote the $k^{th}$ model's predictive density for $y^*$, the combined distribution for $y^*$  is given by, $p(y^*) \propto \prod_{k = 1}^L p_k^{\beta_k(\bm x^*)}(y^*)$.

The function $\beta_k(\bm x^*)$ in the exponent acts as a weight for the $k^{th}$ model's prediction based on the input $\bm x^*$ such that  $\sum_{k = 1}^L \beta_k(\bm x^*) = 1$. 
When each of the $k$ predictions is a Gaussian distribution with mean $\mu_k(\bm x^*)$ and variance $\sigma_k(\bm x^*)$, the combined distribution is also Gaussian. The resulting mean and variance are given by $\mu(\bm x^*) = \sigma^2(\bm x^*)\sum_{k = 1}^L\beta_k(\bm x^*)\sigma^{-2}_k(\bm x^*)\mu_k(\bm x^*)$ and $\sigma^{2}(\bm x^*) = (\sum_{k = 1}^L\beta_k(\bm x^*)\sigma^{-2}_k(\bm x^*))^{-1}$, respectively. For details on computing $\beta_k$, see \cref{appendix:method_detail}.
 
Although we have focused on the combined predictive distribution for the observed $y^*$, in the case of a Gaussian likelihood the same results hold for the combined distribution of the latent function $f^*$ (with the variance terms changed to reflect we are working with the noiseless latent function). For details on combining predictions in $f$ or $y$ space, see \cref{appendix:combining}. 

\subsection{Extracting Information from Intermediate Representations}\label{sec:background_intermediate_reps}
Consider a finite sequence of functions $f_1, ..., f_{L+1}$ such that $f_1:\mathbb{R}^d\rightarrow \mathcal{A}_1$, $f_2:\mathcal{A}_1\rightarrow\mathcal{A}_2$, ...,$f_{L}:\mathcal{A}_{L-1}\rightarrow\mathcal{A}_L$, $f_{L+1}:\mathcal{A}_{L}\rightarrow\mathbb{R}$, where each $\mathcal{A}_i \subset \mathbb{R}^{d_i}$.

The $k^{th}$ \textbf{intermediate representation} of a point $\bm x\in \mathbb{R}^d$ with respect to the sequence $f_1, ..., f_{L+1}$ is defined to be $\bm e_k(\bm x) := (f_k\circ f_{k-1}\circ ... \circ f_1)(\bm x)$. The $k^{th}$ \textbf{intermediate space} $\mathcal{A}_k$ is the image of $(f_k\circ f_{k-1}\circ ... \circ f_1)(\cdot)$.

A \textbf{composite model} $\mathcal{M}$ is the composition $M(\bm x):= (f_{L+1}\circ f_{L-1}\circ ... \circ f_2 \circ f_1)(\bm x)$. A feed-forward network with $L$ layers is an example of a composite model, and the output of each layer in the network, excluding the final layer, results in an intermediate space.

\subsection{Aleatoric and Epistemic Uncertainty}
Consider a mapping $h: \mathcal{X} \rightarrow \mathcal{Y}$ and a tunable model $f_\theta : \mathcal{X} \rightarrow \mathcal{Y}$ aiming to approximate this mapping. Given a dataset $\mathcal{D}_n = \{x_i, y_i\}_{i=1}^n$ generated by $h$, we encounter two types of uncertainties in machine learning.
\textit{Aleatoric uncertainty} originates from the inherent noise in the process $h$, such as variability in data generation, and is irreducible. In contrast, \textit{epistemic uncertainty} stems from the model $f_\theta$, arising due to either model misspecification or insufficient data to accurately replicate $h$. Unlike aleatoric uncertainty, epistemic uncertainty can be reduced by refining the model or expanding $\mathcal{D}_n$. We aim to estimate the total uncertainty in predictions from $f_\theta$, and to distinguish between aleatoric and epistemic components.

\section{RELATED WORK}\label{sec:related_work} 
In this section, we review work on UQ for DNNs, relevant GP approximations, and methods that make use of the internal representations of a neural network. We finish by briefly mentioning the relation of our method to a few other related techniques. 

For UQ with DNNs popular methods include ensembles of DNNs \parencite{lakshminarayanan2017deepensemble} and Bayesian neural networks (BNNs) \parencite{neal2012bayesiandl}. Ensemble methods train multiple DNNs on different data subsets, while BNNs treat network weights as random variables and use Bayesian inference to estimate a posterior distribution for these weights. Predictions in BNNs involve averaging over this posterior, thus accounting for weight uncertainty. Due to the computational complexity, approximations such as sampling or posterior approximation techniques are often employed \parencite{neal2012bayesiandl, ritter2018laplace, chen2014sghmc, gal2016dropout, maddox2019swag}. Since our approach is based on a single network there is potential for integration our work with these existing frameworks.

Deterministic uncertainty quantification (UQ) methods offer an alternative to Bayesian neural networks (BNNs) and ensembles by estimating uncertainty through distances in feature space \parencite{amersfoot2020DUQ, liu2020sngp, amersfoot2021due, mukhoti2023ddu}. This requires the network's penultimate layer, $f_{\bm \theta}(\bm x)$, to be both sensitive to input perturbations and smooth, often enforced via a bi-Lipschitz constraint. A well-structured feature space enables the construction of a distance-aware predictive distribution $p(y|\bm x)$ that performs well for both in-distribution and out-of-distribution (OOD) data. However, deep networks often suffer from feature collapse, where distinct inputs map to nearly identical features, limiting the reliability of uncertainty estimates \parencite{amersfoot2020DUQ}.  

To address feature collapse, deterministic UQ methods modify network training using either two-sided gradient penalties \parencite{amersfoot2020DUQ} or spectral normalization (SN) \parencite{liu2020sngp, amersfoot2021due, mukhoti2023ddu}. Deterministic Uncertainty Quantification (DUQ) enforces a two-sided gradient penalty to train radial basis networks \parencite{amersfoot2020DUQ}. Spectral Normalized Gaussian Process (SNGP) instead applies SN during training and approximates a Gaussian process output layer using random Fourier features (RFF) and a Laplace approximation to the posterior of coefficients \parencite{liu2020sngp, liu2023sngpJMLR}. Similarly, Deterministic Uncertainty Quantification (DUQ) \parencite{mukhoti2023ddu} incorporates SN but uses Gaussian discriminant analysis for OOD detection and network logits for in-distribution uncertainty estimates. Deterministic Uncertainty Estimation (DUE) \parencite{amersfoot2021due} also relies on SN, enforcing a bi-Lipschitz on the feature extractor used in deep kernel learning (DKL) \parencite{wilson2016dkl}. While these methods improve uncertainty estimation, they assume control over network training, preventing their application to pretrained models. 

Our approach, in contrast, builds on the study of neural collapse, a phenomenon closely related to feature collapse. In neural collapse, a classifier maps unique inputs to identical points based on class label. Recently it has been observed that neural collapse typically occurs at an intermediate layer of a network near the output \parencite{rangamani2023intermediatecollapse}. Given this insight, we develop a distance-aware predictive distribution using networks not strictly adhering to bi-Lipschitz smoothness.

Our distance-aware predictive distribution heavily relies on approximating GPs. We briefly review relevant GP approximations. The Vecchia approximation \parencite{vecchia1988estimation,katzfuss2021general} enhances computational efficiency through ordered conditional approximations using nearest neighbors. Inducing-point methods \parencite{hensman2013svigp, titsias2009variational}, on the other hand, introduce a low-rank covariance matrix structure. Ensembling techniques improve scalability by distributing data across independent GPs, with ongoing research \parencite{cohen2020wassersteinBarycenter, rulliere2018nestedgp} dedicated to addressing and overcoming their inherent challenges.

A few notable methods utilize intermediate representations of neural networks. These include deep kernel learning \parencite{wilson2016dkl}, which inputs the penultimate layer's output into a GP without addressing feature collapse---a key aspect for uncertainty quantification as illustrated in our S-curve example (\cref{sec:experiments_intermediate_space_viz}). Deep k-nearest neighbors (DKNN) \parencite{papernot2018deep} assesses test point conformity for uncertainty measurement using intermediate representations and neighbor labels, rather than distance. Differing from the deep GP \parencite{damianou2013deep}, our approach feeds network layer outputs into Vecchia GPs instead of cascading outputs from one GP to another. A particularly related approach uses Gaussian process probes (GPP) \parencite{wang2023gpp}, which define a prior distribution over classifiers that take activations as inputs. Conditioning on activation-label pairs yields a posterior over these classifiers, enabling assessment of both aleatoric and epistemic uncertainty. While similar to our work, GPP work on a single layer at a time. Combining our approach with GPP could extend DVE to classification tasks while allowing GPP to process multiple layers simultaneously.

%% file: contents/03_methods.tex
\section{DEEP VECCHIA ENSEMBLE}\label{sec:model_construction}
We are provided with a dataset $\mathcal{D}$ comprising $n$ input-output pairs, $\mathcal{D}=\{\bm x_i, y_i\}_{i=1}^n$, and a composite model $\mathcal{M}:\mathcal{X}\rightarrow\mathbb{R}$ that is trained to predict $y_i$ given the corresponding $\bm x_i$. Initially, we map the inputs $\{\bm x_i\}_{i=1}^n$ to their respective intermediate representations, $\{\bm e_k(\bm x_i)\}_{i=1}^n$, where $k = 1,...,L$. We generate a collection of conditioning sets, $\mathcal{G}_k$, for each set of intermediate representations, $\{\bm e_k(\bm x_i)\}_{i=1}^n$, by computing ordered nearest neighbors based on the Euclidean distance between the intermediate representations. The resulting tuple for each intermediate space $(\{\bm e_k(\bm x_i), \bm y_i\}_{i=1}^n, \mathcal{G}_k)$ is combined with an automatic relevance determination (ARD) kernel $K_{\bm \theta_k}(\cdot, \cdot)$ of the appropriate dimension to form an ensemble of deep Vecchia GPs.

\begin{figure}
    \centering
    \includegraphics[width =1.\linewidth]{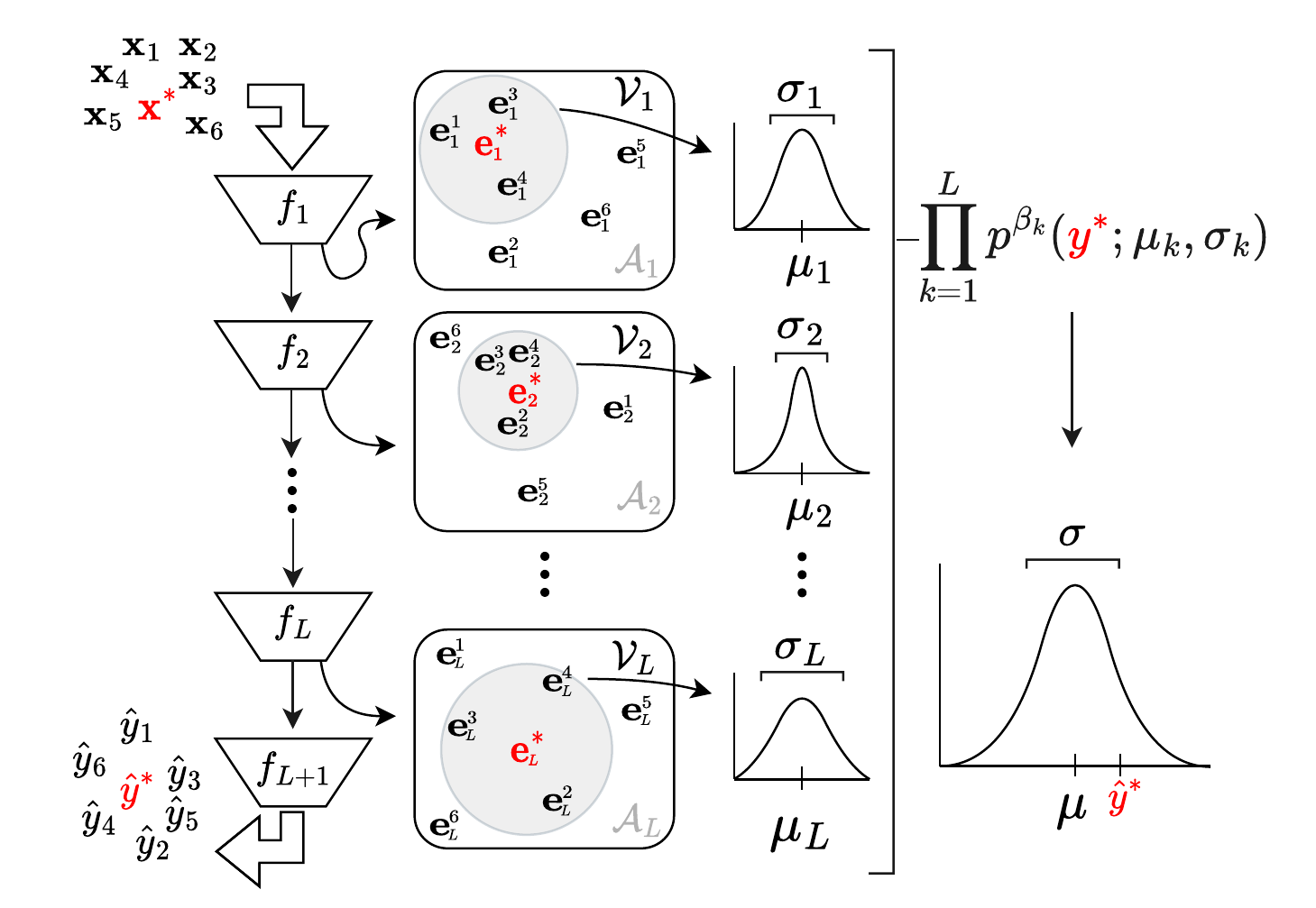}
    \caption{\textbf{The DVE pipeline is conceptually simple.} The network maps inputs $\bm x_1, \bm x_2, \bm x_3, \ldots, \bm x_n$, to intermediate spaces, where we compute nearest neighbors. For example, in $\mathcal{A}_1$, the neighbors for $\textcolor{red}{x^*}$ are $\bm e_1^1 , \bm e_1^3, \bm e_1^4$ (for simplicity, we define $\bm e_k^i=\bm e_k(\bm x_i)$). For each layer, we define a Vecchia GP, denoted by $\mathcal{V}_1, \ldots, \mathcal{V}_L$, which estimates a distribution for the response $\textcolor{red}{y^*}$. Estimates are combined in a product-of-experts fashion to yield a single distribution parameterized by $\hat{\mu}$ and $\hat{\sigma}$, which improves upon the network's point prediction $\textcolor{red}{\hat{y}^*}$.}
    \label{fig:model_summary}
\end{figure}

At test time, the test point will be mapped to its intermediate spaces and each Vecchia GP will provide a predictive mean and variance. The predictions are then combined via a product of experts to form a single mean and covariance estimate. \cref{fig:model_summary} provides an overview of the proposed model.

\subsection{Dataset Extraction from a Neural Network}\label{sec:model_construction_layers}
To begin, we describe how we extract a sequence of datasets from a pretrained neural network and the data it was trained on. Suppose the first layer of our pretrained network is of the form $f_1(\bm x) = \sigma (\bm x^T \bm W^{(1)} + \bm b^{(1)})$, with, $\bm W^{(1)} \in \mathbb{R}^{d \times d_1}$, $\bm b_1 \in \mathbb{R}^{d_1}$ and $\sigma_1:\mathbb{R}^{d_1}\rightarrow\mathbb{R}^{d_1}$. Then $\bm x$'s first intermediate representation, $\bm e_1(\bm x) := f_1(\bm x)$, will lie in a subset of $\mathbb{R}^{d_1}$, that is  $\bm e_1(\bm x) \in \mathcal{A}_1\subset\mathbb{R}^{d_1}$. Similarly, $\bm x$'s second intermediate representation is given by, $\bm e_2(\bm x) := (f_2\circ f_1)(\bm x)$, where $f_2(\bm e_1(\bm x)) = \sigma (\bm e_1(\bm x) \bm W^{(2)} + \bm b^{(2)})$ with $\bm W^{(2)} \in \mathbb{R}^{d_1 \times d_2}$, $\bm b_2 \in \mathbb{R}^{d_2}$ and $\sigma_2:\mathbb{R}^{d_2}\rightarrow\mathbb{R}^{d_2}$. We can define the remaining $L-2$ intermediate representations in a similar way. 

By combining the intermediate representations with the responses, we can generate $L$ datasets $\{\mathcal{D}_k\}_{k=1}^L$, where $\mathcal{D}_k = \{\bm e_k(\bm x_i), y_i\}_{i=1}^n$. We assume the same random ordering of observations for all $L$ datasets, so $y_i$ will always refer to the same response for every $\mathcal{D}_k$. 

\begin{figure*}
    \centering
    \includegraphics[width =.9\linewidth]{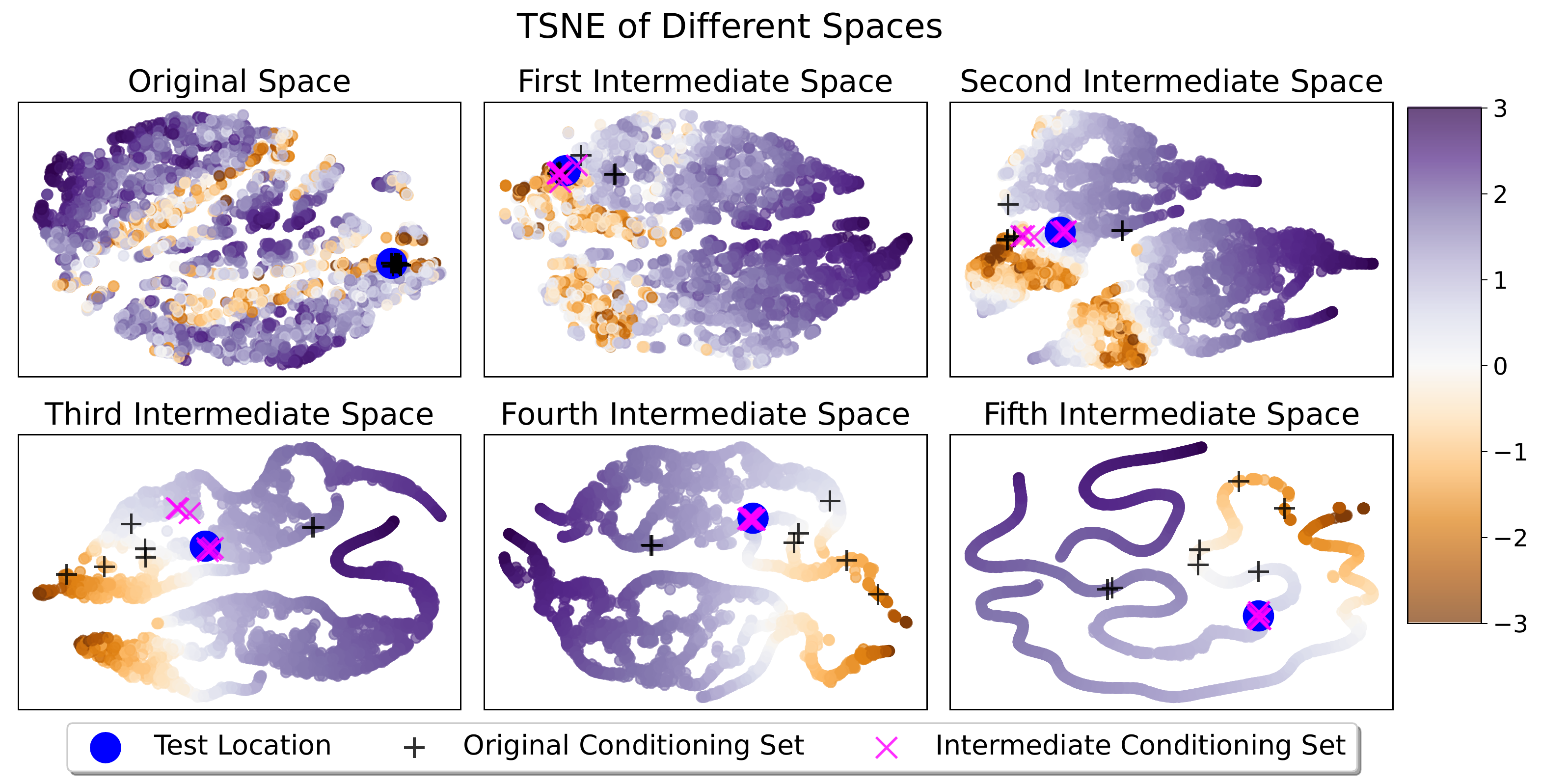}
    \caption{\textbf{Sensitivity and smoothness vary by layer}. The 2D TSNE projection of input points in the original space of the bike UCI dataset (top left panel) and intermediate spaces (remaining five panels) with the color denoting the response value. The blue dot is a fixed test point that has been propagated through the network. The black crosses denote the ordered nearest neighbors of the blue dot in the original input space, and their corresponding intermediate representations. The magenta crosses denote the ordered points nearest to the blue point within each intermediate space.}
    \label{fig:experiments_intermediate_space_viz_bike}
\end{figure*}

An example of generating $L$ datasets is visualized in \cref{fig:experiments_intermediate_space_viz_bike} for the bike UCI dataset. Each of the panels contains one dataset to which we fit a Vecchia GP. 

\textbf{Why do we need intermediate layers?} Any last-layer method \parencite{amersfoot2020DUQ, liu2020sngp, mukhoti2023ddu} needs to modify training to address feature collapse \parencite{amersfoot2020DUQ}. With multiple layers, distinct points in the input space will be distinct for at least one layer, as neural collapse tends to occur later in the network \parencite{rangamani2023intermediatecollapse}. The heterogeneity of inputs and corresponding responses for different layers also benefits prediction and UQ.

\subsection{Conditioning-Set Selection}\label{sec:background_cond_sets}
\cref{fig:experiments_intermediate_space_viz_bike} suggests that the function $\bm e_k(\bm x) \mapsto y$ is smooth for some value of $k$; therefore, we choose to estimate each $\bm e_k(\bm x) \mapsto y$ using a GP. For computational purposes, we approximate each of the GPs using Vecchia with nearest neighbors computed in the corresponding intermediate space. Doing so gives rise to several conditioning sets for each $\bm x_i$. 

Letting $g_k(i)$ denote the conditioning set for observation $i$ from intermediate space $k$, we consider the \textbf{collection of conditioning sets}, given by $\mathcal{G}_k := \{g_k(1), ..., g_k(n)\}$. An $L+1$ layer network will have $L$ collections of conditioning sets, $\mathcal{G}_1, ..., \mathcal{G}_L$. 

By using several collections of conditioning sets, we utilize the unique information present in the panels titled `First Intermediate Space' through `Fifth Intermediate Space' in \cref{fig:experiments_intermediate_space_viz_bike}. This approach can lead to better predictions compared to using just one collection of conditioning sets.

\textbf{Why not use inducing-point GPs?} Our primary contribution is the use of intermediate-layer outputs for deterministic UQ, and inducing-point GPs could be used as a replacement for the Vecchia approximation in this situation. 

\textbf{What are the benefits of using the Vecchia approximation in particular?} The primary benefits are interpretability and distinguishing between aleatoric and epistemic uncertainty. Interpretability is achieved by examining test-point conditioning sets, revealing the training examples deemed relevant by the network. Using nearest neighbors also allows for precise UQ: aleatoric uncertainty, reflected in the Gaussian likelihood's noise term, indicates data variability, while epistemic uncertainty is gauged by the test point's proximity to training instances. Notably, in ambiguous cases---such as \cref{sec:inspecting_cond_sets}---the conditioning sets pinpoint the source of uncertainty, a nuance that inducing points lack, as they cannot differentiate whether uncertainty stems from data sparsity, proximity to inducing inputs, or conflicting data interpretations.

\subsection{Ensemble Building from Conditioning Sets}\label{sec:background_combining_gps}
Each $\mathcal{G}_k$ represents a hypothesis about the conditional independence structure of the data. For instance, with time series data produced by an AR(2) process, the $i^{th}$ observation should depend on the two preceding observations in time, while for an AR(1) process, the conditioning set is limited to the immediately preceding data point. However, in general, it is unclear which $\mathcal{G}_k$ is most appropriate.

Since we do not know which $\mathcal{G}_k$ is most suitable, we propose an ensemble of Vecchia GPs, each of which is based on one of the collections of conditioning sets. With each Vecchia GP independently fit to its corresponding dataset (e.g. $\mathcal{D}_k$), we combine the predictions from each member of the ensemble in a product-of-experts fashion. We combine our predictions in $y$-space; details on this choice are given in \cref{appendix:combining}. Additional details on training and prediction are provided in  \cref{appendix:method_detail}.

%% file: contents/04_applications.tex
\section{APPLICATION TO PRETRAINED MODELS}\label{sec:applications}
We conduct experiments using pretrained networks across various tasks, comparing DVE to the original network and adjusting the choice of baselines depending on the task requirements. For the chemical property prediction task, we compare DVE with the base model and last-layer Gaussian process (GP) methods, as retraining the networks for these tasks is not feasible. In contrast, for the UCI regression tasks, we benchmark against ensembles, Monte Carlo (MC) dropout, and deep kernel learning, since retraining the networks is straightforward in this case. We measure performance using root-mean-square error (RMSE) and negative-log likelihood (NLL).

\subsection{UCI Benchmarks}\label{sec:applications_uci}
We train a feed-forward neural network on three train-validation-test splits of UCI regression tasks. After training, we construct the deep Vecchia ensemble (DVE) as in \cref{sec:model_construction} and we make predictions for the test set using the method in \cref{appendix:method_detail}. Additionally, we train the stochastic variational inference GP (SVI-GP) of \textcite{hensman2013svigp}, the deep Gaussian process (Deep GP) of \parencite{damianou2013deep}, and the scaled Vecchia GP (Vecchia) of \parencite{katzfuss2022scaled} separately on the training data for prediction on the test set.

\cref{tab:application_uci} compares NLL (left column under each dataset) and RMSE (right column under each dataset) for different methods on UCI datasets. The bold value in each column is the best performing (lower is better). The value in parentheses is the sample standard deviation of the mean of the metric over the three different data splits. The last line of the table specifies the sample size (n) and dimension (d) for each dataset. 

We can see that either DVE or the deep GP have the best NLL across the datasets, but the DVE is easier to work with given it is built on a pretrained network. Compared to the neural network, DVE is able to provide uncertainty estimates that are better than existing GP approximations, and DVE even improves upon the neural network’s point predictions (in terms of RMSE) for three of the five datasets. For the full version of \cref{tab:application_uci} and additional baseline comparisons using an alternative network see \cref{appendix:uci_details}.

\begingroup
\setlength{\tabcolsep}{2pt}
\begin{table*}[t]
    \scriptsize
    \caption{\textbf{DVE is consistently competitive with baselines}. NLL $\downarrow$ (left columns) and RMSE $\downarrow$ (right columns) on UCI benchmarks.}
    \label{tab:application_uci}
    \centering
    \begin{tabular}{lllllllllll}
    \toprule
    \multicolumn{2}{c}{Method}& \multicolumn{2}{c}{Elevators}  & \multicolumn{2}{c}{KEGG} & \multicolumn{2}{c}{bike}  & \multicolumn{2}{c}{Protein}\\
    \cmidrule(lr){1-2}\cmidrule(lr){3-4}\cmidrule(lr){5-6}\cmidrule(lr){7-8}\cmidrule(lr){9-10}
    
    \multicolumn{2}{c}{DVE} &

     \textbf{0.350 (0.010)} & 
     0.345 (0.004) &
     -1.021 (0.030) & 0.087 (0.002) &
     \textbf{-3.484 (0.054)} & \textbf{0.002 (0.000)} &
     \textbf{0.646 (0.029)} & \textbf{0.494 (0.014)}
    \\
    \multicolumn{2}{c}{Neural Net} & 

     N/A & \textbf{0.340 (0.003)} &
     N/A & \textbf{0.085 (0.002)} &
     N/A & 0.006 (0.000) & 
     N/A & 0.514 (0.042)
    \\
    \multicolumn{2}{c}{Vecchia} & 

     0.525 (0.015) & 0.414 (0.006) & 
     -0.987 (0.026) & 0.090 (0.002) & 
     -0.239 (0.010) & 0.183 (0.001) & 
     0.851 (0.003) & 0.571 (0.002) & 
    \\
    \multicolumn{2}{c}{SVI-GP} & 

     0.455 (0.017)& 0.381 (0.007) & 
     -0.912 (0.033) & 0.098 (0.003) & 
     -1.140 (0.047)& 0.073 (0.004) & 
     1.083 (0.001) & 0.713 (0.001) & 
    \\

    \multicolumn{2}{c}{Deep GP (3)} & 

     0.366 (0.029) & 0.346 (0.011) & 
     \textbf{-1.036 (0.036)} & 0.086 (0.004) & 
     -2.806 (0.100) & 0.004 (0.001) & 
         0.938 (0.025) & 0.597 (0.018) & 
    \\

    \cmidrule(lr){2-11}
    \multicolumn{2}{c}{($n$,$d$)} & \multicolumn{2}{c}{(16.6K,18)}  & \multicolumn{2}{c}{(48.8K,20)} & \multicolumn{2}{c}{(17.4K,17)}  & \multicolumn{2}{c}{(45.7K,9)}
    \\
    \bottomrule
    \end{tabular}
\end{table*}
\endgroup

\subsection{Chemical Property Prediction}\label{sec:chemical_props}
We applied our method to the Chemformer, as described by \textcite{irwin2022chemformer}, initially pretrained on a 100 million SMILES string subset from the ZINC-15 dataset \parencite{sterling2015zinc} (details in Section 2.1 of \textcite{irwin2022chemformer}). This model was further fine-tuned on three MoleculeNet property prediction datasets—ESOL, FreeSolvation, and Lipophilicity—as used in \textcite{irwin2022chemformer}. Details on data processing, fine-tuning, and deep Vecchia ensemble construction are provided in \cref{appendix:chemical_props}.

Table 3 presents RMSE and NLL results. 'Chemformer' refers to the original fine-tuned network. 'DVE (m = n)' denotes the deep Vecchia ensemble with an exact GP (which is equivalent to a Vecchia approximation with $m=n$), while 'DVE (m = 200)' and 'DVE (m = 400)' use the Vecchia approximation with 200 and 400 nearest neighbors, respectively. Both the exact and Vecchia GP use the exact GP's hyperparameters to focus on the impact of a reduced conditioning set.

In all three datasets, both the exact and approximate GP in DVE reduced RMSE compared to the network alone. The exact GP showed only marginal RMSE improvement over the Vecchia approximation, which generally yielded better NLL. For the smaller ESOL and FreeSolvation datasets, we used exact nearest neighbors, while for the larger Lipophilicity dataset, we employed approximate nearest neighbors with product quantization \cite{jegou2010product, johnson2017similarity}. The choice between exact and approximate nearest neighbors did not significantly impact the results.

\begin{table}[t]
    \scriptsize
    \caption{\textbf{DVE improves performance of Chemformer}. RMSE and NLL for property prediction.}
    \label{tab:chem_prop}
    \centering
    \begin{tabular}{llll}
    \toprule
    Dataset & Model & RMSE $\downarrow$ & NLL $\downarrow$\\
    \cmidrule(lr){1-1}\cmidrule(lr){2-2}\cmidrule(lr){3-3}\cmidrule(lr){4-4}
    &Chemformer & 0.57 (0.03) & N/A\\
    ESOL &DVE ($m = n$)& 0.35 (0.01) &  -2.02 (0.16)\\
    &DVE ($m = 200$) & \textbf{0.35} (0.01) & \textbf{-2.21} (0.05)\\
    \cmidrule(lr){2-4}
    &Chemformer & 2.23 (0.15) & N/A\\
    FreeSolvation &DVE ($m = n$) & \textbf{1.30} (0.09) &  -1.45 (0.10)\\
    &DVE ($m = 200$) & 1.32 (0.09) &  \textbf{-1.49} (0.10)\\
    \cmidrule(lr){2-4}
    &Chemformer & 0.73 (0.02) & N/A\\
     Lipophilicity &DVE ($m = n$)& \textbf{0.60} (0.01) &  -0.60 (0.11)\\
    &DVE ($m = 400$) & 0.61 (0.01)& \textbf{-0.63} (0.09)\\
    \bottomrule
    \end{tabular}
\end{table}

%% file: contents/05_experiments.tex
\section{EXPERIMENTS}\label{sec:experiments}
In this section we explore DVE properties with the goal of understanding how our procedure works. We begin by looking at the intermediate representations of a network and then we check what the conditioning sets of a neural network look like for correct and incorrect predictions.
\subsection{Exploring Intermediate Spaces}\label{sec:experiments_intermediate_space_viz}
\begin{figure}
    \centering
    \includegraphics[width =1.0\linewidth]{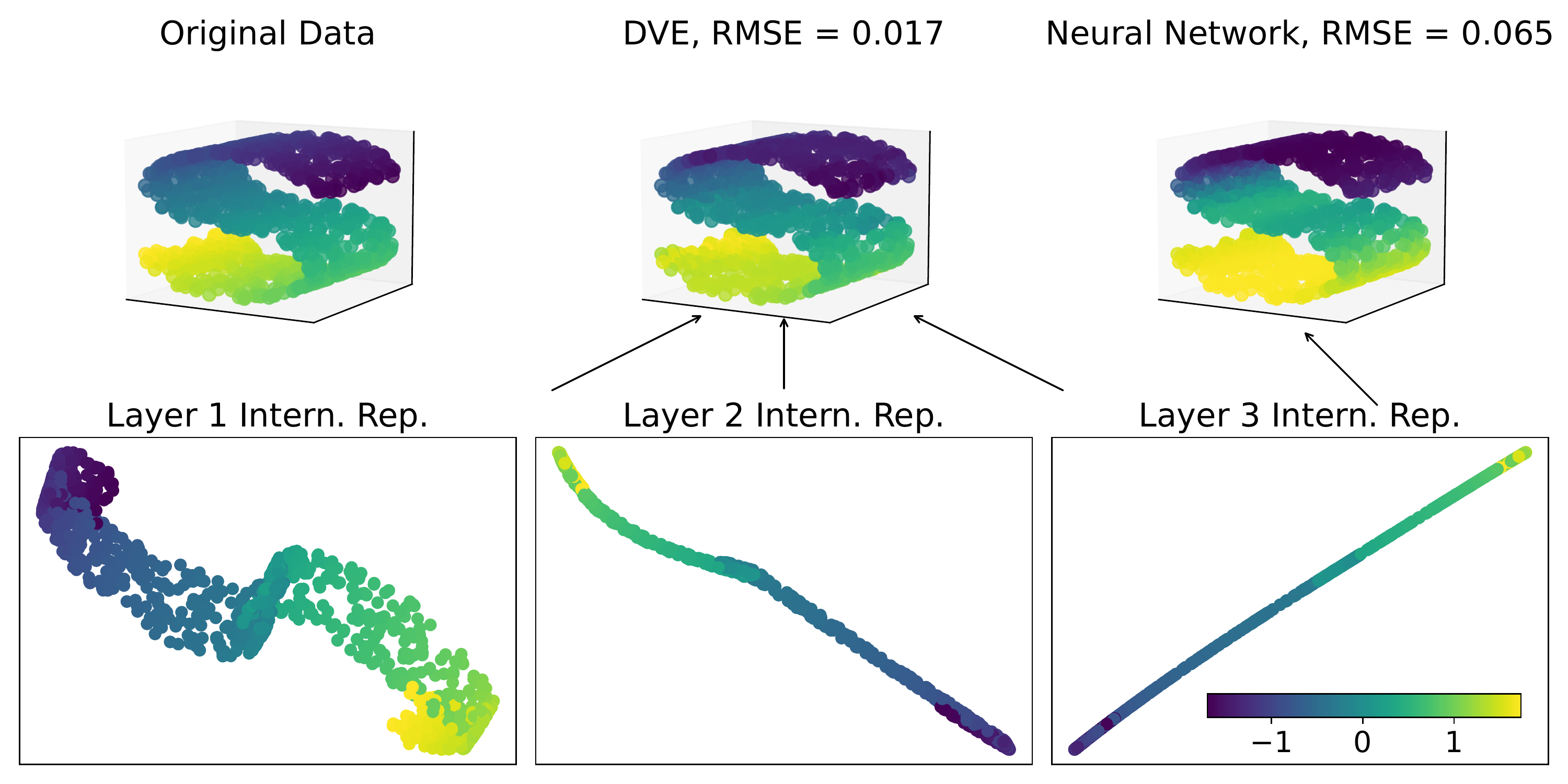}
    \caption{\textbf{Ensembling layers can outperform last layer in terms of MSE}. Top left: Data from the S-curve dataset. Top middle: Predictions generated by DVE. Top right: Predictions from the neural network. In all panels, the color denotes the response value. Arrows emphasize that DVE predictions use all three layers, whereas the final network prediction uses only the last layer.}
    \label{fig:s_curve}
\end{figure}

To investigate how DVE uses different layers we work with a feed-forward network fit to samples from the S-curve dataset \parencite{tenenbaum2000scurve}. \cref{fig:s_curve} shows results and a basic description is given in the caption. For additional details, see \cref{appendix:s_curve}. After the first layer the green and yellow points are separated but these points collide by the final layer. DVE preserves this distinction, unlike the network. The DVE predictions closely match the original dataset, with a lower MSE of 0.018 compared to the neural network's 0.065. 

Similar results hold for different networks. For example, with four hidden layers with (10, 5, 2, 2) units, the MSE dropped from 1.8e-1 to 4.2e-5. Changing the number of units per layer to (16, 8, 4, 4) showed the MSE drop from 7e-3 to 3.9e-5. 

\subsection{Interpreting Internal Representations}
DVE can be seen from two perspectives: ensembling Vecchia GPs, and a method for UQ in DNNs. Below we relate both view points to the results in this section. 

From the Vecchia ensemble perspective, we can see the intermediate representations in \cref{fig:s_curve} as providing three different distance metrics that can be used to compute possibly unique conditioning sets. The requirements to make this connection are formalized in \cref{prop:sequences}. 

\begin{prop}\label{prop:sequences}
Consider a sequence of injective functions $f_1, .., f_L, f_{L+1}$ and a sequence of metric spaces $\{(M_i, d_i)\}_{i=1}^{L+1}$ such that $f_i: \mathcal{M}_i \rightarrow \mathcal{M}_{i+1}$ for $i = 1, ..., L$. Then $\{(M_i, \tilde{d}_i)\}_{i=2}^{L+1}$ defines a sequence of metric spaces, where $\tilde{d}_i(x,y) := d_i(\bm e_i(\bm x), \bm e_i(\bm x'))$ for any $\bm x,\bm x' \in M_1$, with $\bm e_i(\bm x):=f_i \circ f_{i-1}\circ ... \circ f_1(\bm x)$ . 
\end{prop}

Requiring the functions be injective ensures unique internal representations for unique input points, but continuity is not required. Discontinuous layers are useful when modeling functions with high Lipschitz constants or discontinuities. In such cases, predicting values using Euclidean nearest neighbors in $\mathbb{R}^d$ can be inaccurate, but the function may be continuous with respect to a different distance metric on $\mathbb{R}^d$. The difference in continuity is reminiscent of what we observed in \cref{fig:experiments_intermediate_space_viz_bike}, where the function became progressively smoother for later layers. Using multiple spaces allows for competing hypothesis about what distance metric is best.

From the UQ perspective, we can see the sequence of distance metrics as measuring how far a test point is away from instances in the training data. Of course, if a test point is close to existing training data with respect every distance metric, then the ensemble of Vecchia GPs will provide an estimate with small uncertainty. For data far from the training set, the internal representations will not match those of the training data and the prediction will be close to the GP prior. 

\textbf{If a network is experiencing feature collapse, will the sequence of functions be injective?} In short, no, the entire sequence of functions may not be injective. However, there is likely a subsequence of the original sequence that remains injective. The DVE can handle non-injective functions effectively because of the posterior variance weighting we use. Specifically, when functions are non-injective, the posterior variance is dominated by output-scale noise, which accommodates the differing function values for unique inputs. As a result, the DVE assigns less weight to the non-injective functions, allowing it to approximate the two perspectives mentioned earlier. This property, along with the ablation studies in \cref{appendix:combining_gps_experiment}, highlights the strong appeal of posterior variance weighting.

\subsection{Inspecting Conditioning Sets}\label{sec:inspecting_cond_sets}
In this experiment, we train a feed-forward network on MNIST and generate embeddings from train data using two hidden layers. Then we select two test set instances—one correctly labeled and one incorrectly labeled by the network—and analyzed their embeddings. 

\begin{figure}[t]
    \centering
    \includegraphics[width = 1.0\linewidth]{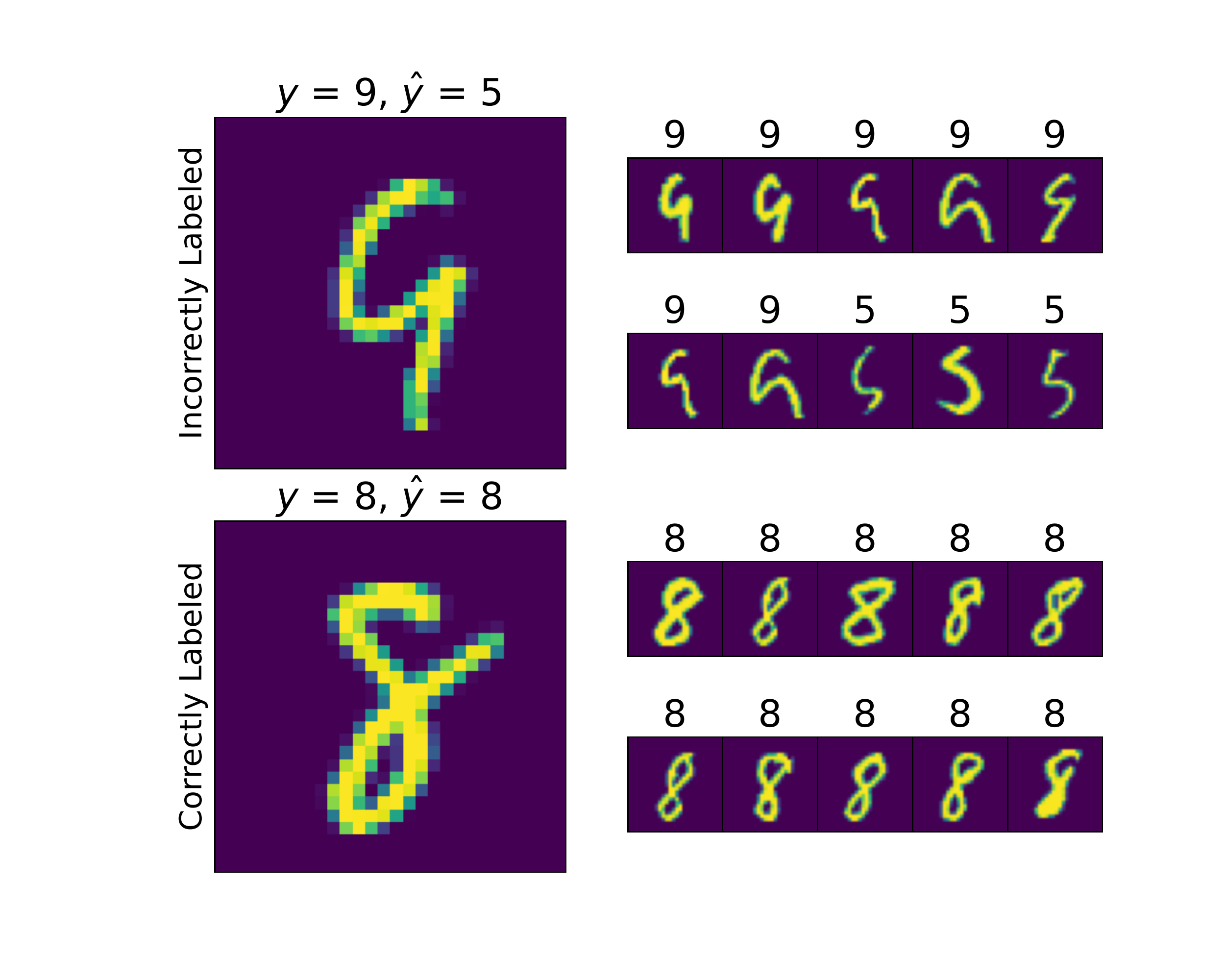}
    \caption{\textbf{Layerwise conditioning sets relate to aleatoric uncertainty}. Top left: A test image of a digit 9 incorrectly classified as a 5 by the neural network. Top right: Nearest-neighbor images for the misclassified 9 at the third (top row) and fourth (bottom row) layers of the network, with the top row showing neighbors of the correct class (9) and the bottom row indicating emergence of the incorrect class (5). Bottom half: For a correctly labeled test image of an 8 (bottom left), the nearest-neighbor images at both the third (top row) and fourth (bottom row) layers (bottom right) consistently show the correct class (8).}
    \label{fig:cond_sets}
\end{figure}

As \cref{fig:cond_sets} illustrates, the incorrectly labeled example's nearest neighbors in the first intermediate space are all nines, while in the second space, they are a mix of nines and fives. Conversely, the correctly labeled example's nearest neighbors are consistently eights in both spaces. The incorrectly labeled example is notably more distant from its nearest neighbors, being twice as far in the first embedding and five times in the second. This highlights how the network differently processes correctly and incorrectly classified examples, a phenomenon leveraged in our method. The ensemble mean acts as a weighted average of training labels, indicating layer agreement or providing a nuanced prediction when layers disagree.

%% file: contents/06_conclusion.tex
\section{CONCLUSION}\label{sec:conclusion}
We introduced DVE for deterministic UQ with pretrained neural networks that does not require altering network training. The effectiveness of DVE was demonstrated across various tasks, including benchmark regression and chemical property prediction. Additionally, analyses of internal representations and conditioning sets offer valuable insights into the method's functionality.

Unlike existing deterministic methods \parencite{liu2020sngp, amersfoot2020DUQ, mukhoti2023ddu}, our method works with any existing neural network without changing training, expanding the scope of deterministic UQ methods. The proposed method is relevant for latent-space Bayesian optimization \parencite{gomez2018lsbo}, where it could streamline the process by eliminating the need for additional latent-space structuring, as seen in \textcite{grosnit2021metricLearning}. 

Our model has three main limitations. Firstly, it requires the original training data, which may not always be accessible. A potential workaround is to fit the network alongside an ensemble of inducing-point GPs or use a smaller validation set for the deep Vecchia ensemble. Secondly, our approach combines predictions in $y$-space, limiting us to Gaussian likelihoods. This can be addressed by combining predictions in $f$-space and establishing a method to merge the noise terms from each Vecchia GP's likelihoods (see \cref{appendix:f_vs_y}). Finally, we utilize a DNN with fixed weights, whereas many BNN methods treat weights as random variables. Integrating DVE into BNNs could involve representing intermediate representations as distributions, using an error-in-variables Vecchia GP that combines \textcite{bachoc2023gaussian}'s kernel with \textcite{kang2023correlation}'s correlation strategy.

%% file: contents/07_appendix.tex
\section{METHOD DETAILS}\label{appendix:method_detail}
In this section, we provide details on the Deep Vecchia Ensemble (DVE) method proposed in \cref{sec:model_construction}. Specifically, \cref{appendix:covar} provides details on our choice of covariance function, \cref{appendix:training} outlines the process of fitting the DVE, \cref{appendix:prediction} details the steps for predicting with the DVE, \cref{appendix:complexity} discusses the complexity of our proposed method, \cref{appendix:prop_1_proof} presents the proof of \cref{prop:sequences}, and \cref{appendix:combining} discusses how to combine the predictions from the layer-wise GPs of DVE.

\subsection{Covariance Function}\label{appendix:covar}
For each GP, we use an ARD Mat\'ern 5/2 kernel, introducing $(d + 1)$ parameters from the kernel that must be estimated. We select the Mat\'ern kernel because it has a natural connection to the screening effect \parencite{stein2011screening}, where spatial predictions primarily depend on observations closest to the prediction location. Briefly, the Mat\'ern covariance function satisfies conditions that ensure an asymptotic screening effect holds \parencite{stein2011screening, kang2024asymptotic}, which is not true for the squared-exponential kernel \parencite{stein2015noscreening}. While empirical evidence suggests the squared-exponential kernel performs well with the Vecchia approximation \parencite{kang2024asymptotic}, we chose the Mat\'ern covariance due to its strong theoretical connection to the screening effect.

\subsection{Training}\label{appendix:training}
For an $L+1$-layer network trained on a dataset with $N$ observations, we begin by generating $L$ datasets, where the inputs are the intermediate representations of the training inputs $\bm{x}$, and the responses are the same responses from the training data. That is, for each $k = 1, \dots, L$, we have $\mathcal{D}_k =\{\bm{e}_k(\bm{x}i), y_i\}{i=1}^{N}$. We will then independently fit a Vecchia Gaussian process to each dataset, $\mathcal{D}_k$.

This implies that for an $L+1$-layer network, we obtain an ensemble of $L$ Vecchia GPs, each with its own set of parameters $\bm{\theta}_k = (\sigma_k^2, \bm{\lambda}_k, \tau_k)$, where $\sigma_k^2$ denotes the output variance for the kernel, $\bm{\lambda}_k \in \mathbb{R}_{+}^{d_k}$ represents the lengthscales for the ARD kernel, and $\tau_k^2$ is the variance in the Gaussian likelihood. To estimate $\bm{\theta}_k$ for each Vecchia GP, we independently apply mini-batch gradient descent within type-II maximum likelihood, optimizing the Vecchia marginal log-likelihood using the corresponding dataset $\mathcal{D}_k$
\begin{align}\label{eqn:full_likelihood}
\hat{\bm{\theta}}_k &= \underset{\bm{\theta} \in \Theta}{\arg\max} \sum_{i=1}^{n} \log p_{\bm{\theta}}(y_i | \bm{y}_{g_k(i)}) \\
&=\underset{\bm{\theta} \in \Theta}{\arg\max} \sum_{i=1}^{n} \log \mathcal{N}\left( y_i \mid \mu^{(k)}_i, \sigma^{(k)}_i + \tau_k^2 I \right).
\end{align}
The mean $\mu^{(k)}_i$ and the variance $\Sigma^{(k)}_i$ are given by,
\begin{align*}
    \mu_i^{(k)} &= \bm{K}_{i, \bm g} \left( \bm{K}_{\bm g, \bm g} + \tau_k^2 I \right)^{-1} \bm{y}_{g_k(i)} \\
    \sigma_i^{(k)} &= \bm{K}_{i, i} - \bm{K}_{i, \bm g} \left( \bm{K}_{\bm g, \bm g} + \tau_k^2 I \right)^{-1} \bm{K}_{\bm g, i},
\end{align*}
with $\bm{K}_{i, \bm g}$ denoting the covariance between $\bm e_k(\bm x_i)$ and $\bm{E}_{g_k(i)}$ where $\bm{E}_{g_k(i)} = \{\bm{e}_k(\bm{x}_j)\}_{j \in g_k(i)}$. 

The conditional distribution $p_{\bm \theta_k}(y_i|\bm y_{g_k(i)})$ can be seen as using nearby observation, $\bm y_{g_k(i)}$, to predict $y_i$. This conditional distribution depends on the parameters $\bm \theta_k$, the intermediate representation, and the conditioning set. This model allows us to effectively capture the unique complexities present in each intermediate space of the neural network. Since we are optimizing the layer-wise Vecchia GPs independently, we ignore the hierarchical structure of the intermediate representations. Training the models independently is fundamentally different from a deep Gaussian process \parencite{damianou2013deep} as we do not feed the outputs of one GP into the next GP. However, by performing our training in this way, we can more easily work with existing pretrained models. Therefore, our procedure can be viewed as augmenting both the Vecchia GP and the pretrained model.

\subsection{Prediction}\label{appendix:prediction}
For a test point $\bm x^*$, we compute the intermediate representation, $\bm e_k^*:= \bm e_k(\bm x^*)$, and find the $m$ nearest neighbors of $\bm e_k^*$ in the training dataset $\mathcal{D}_k$. We then compute the posterior mean $\mu_k^*$ and variance $\sigma_k^*$ of $f^*$, where $y^* | f^* \sim \mathcal{N}(f^*, \sigma_k^2)$. We repeat this for each of the $L$ layers. We then combine the $L$ estimates into a single estimate as given below:
\begin{align}\label{eqn:model_prediction_combined}
    \mu(\bm x^*) &= \sigma^2(\bm x^*) \sum_{k = 1}^L \beta_k(\bm e_k^*) (\sigma_k^*)^{-2} \mu_k^* \\
    \sigma^2(\bm x^*) &= \sum_{k = 1}^L\beta_k(\bm e_k^*) (\sigma_k^*)^{-2}
\end{align}
We chose this combining strategy to account for prediction uncertainty. Specifically, our approach implicitly considers the degree of agreement between the collections of conditioning sets for $\bm{x_i}$ at each layer. If there is high agreement across the layers, we expect the test point is well within the support of our training data and can make confident predictions. In contrast, if there is little agreement, we infer that $\bm{x_i}$ is a relatively novel observation and our predictive uncertainty should reflect this. The relative closeness of the test point and its conditioning set within each intermediate space also influences our confidence in the predictions. For instance, if our model has seen many data points similar to $\bm{x_i}$ at every layer, we would be more confident in our predictions than if our model had never encountered such points, even if the conditioning sets are consistent across all layers. Our product of experts combining strategy captures these behaviors effectively.

As mentioned in \cref{sec:background_gp_ensembles}, we have the choice to combine the predictions in either $f$-space or $y$-space, and we investigate the performance of both approaches in \cref{appendix:combining}. We also investigate the impact of different ways of computing the $\beta_k$'s in \cref{appendix:combining}. In short, we recommend combining methods in $y$-space using the differential entropy to compute the $\beta_k$'s. 

\subsubsection{Layerwise Moments}
Computing layerwise moments is the same for each layer so we focus on a general layer, $k$. The conditional mean and conditional variance for layer $k$ for test point $j$ is given by 
\begin{align*}
    \mu_i^{(k)} &= \bm{K}_{i, \bm g} \left( \bm{K}_{\bm g, \bm g} + \tau_k^2 I \right)^{-1} \bm{y}_{g_k(i)} \\
    \sigma_i^{(k)} &= \bm{K}_{i, i} - \bm{K}_{i, \bm g} \left( \bm{K}_{\bm g, \bm g} + \tau_k^2 I \right)^{-1} \bm{K}_{\bm g, i},
\end{align*}
again with $\bm{K}_{i, \bm g}$ denoting the covariance between $\bm e_k(\bm x_i)$ and $\bm{E}_{g_k(i)}$ where $\bm{E}_{g_k(i)} = \{\bm{e}_k(\bm{x}_j)\}_{j \in g_k(i)}$.

\subsection{Computational Complexity}\label{appendix:complexity}
Below we detail the runtime complexity of training and prediction using the DVE. We let $m$ denote the conditioning set size and $L$ denote the number of layers being ensembled. 

\subsubsection{Training}
Since our approach trains the GP for each layer independently, it inherits the favorable complexity characteristics of the original Vecchia approximation. After computing the nearest neighbors for the training sets using an approximate nearest neighbor (ANN) method, the complexity for each training step is $\mathcal{O}(n_b m^3)$ per layer, where $n_b$ is the number of training mini-batches and $m$ is the size of the conditioning set. For the entire ensemble, the combined complexity for each training step is, $\mathcal{O}(Ln_bm^3)$, where $L$ is the number of layers used by the ensemble. To optimize efficiency, we precompute and store the intermediate representations for each layer on disk, allowing the base network to be queried only once per training point.

\subsubsection{Prediction}
For prediction, we pass a test point through the network once to compute its intermediate representations and then use ANN search at each layer to identify the conditioning sets. The prediction complexity after finding the approximate nearest neighbors is similarly $\mathcal{O}(L n_pm^3)$, where $n_p$ is the number of test points.

\subsubsection{Nearest Neighbors}
The complexity associated with nearest neighbors can be adjusted by tuning the hyperparameters of the ANN algorithm so we excluded it when discussing the training and test complexity. Specifically, we use an Inverted File Index (IVF) approach for computing nearest neighbors, where the data are divided into $N_{\text{list}}$ clusters. Queries involve searching the nearest $N_{\text{probe}}$ clusters to compute nearest neighbors. Increasing $N_{\text{list}}$ improves speed at the cost of accuracy, while increasing $N_{\text{probe}}$ enhances accuracy but reduces speed. This trade-off allows control over the complexity of the nearest neighbor computation. We have observed that large $N_{\text{list}}$ values and small $N_{\text{probe}}$ values can still produce high-quality results. For a more thorough analysis of ANN's impact on Vecchia GPs, we refer to \textcite{jimenez2022scalable}.

\subsection{Proof of Proposition 1}\label{appendix:prop_1_proof}
Let $f:M_1\rightarrow M_2$ be an injective function where $(M_1, d_1)$ and $(M_2,d_2)$ denote metric spaces. Now define $\tilde{d}_1(\bm x, \bm x') := d_2( f(\bm x), f(\bm x'))$ and let $\bm x, \bm x', \bm z \in M_1$.

Note that $\tilde{d}_1(\bm x, \bm x') = d_2( f(\bm x), f(\bm x'))=d_2(f(\bm x'), f(\bm x))=\tilde{d}_1(\bm x', \bm x)$.  

Additionally, $\tilde{d}_1(\bm x, \bm x') = d_2( f(\bm x), f(\bm x')) \geq 0$ for any $\bm x, \bm x' \in M_1$. 

Now suppose $\tilde{d}_1(\bm x', \bm x) = 0 \Leftrightarrow  d_2(f(\bm x'),  f(\bm x)) = 0$  $\Leftrightarrow f(\bm x') =  f(\bm x) \Leftrightarrow \bm x' = \bm x$. 

Finally, observe $\tilde{d}_1(\bm x, \bm x') = d_2( f(\bm x), f(\bm x')) \leq d_2( f(\bm x), f(\bm z)) + d_2( f(\bm z), f(\bm x')) = \tilde{d}_1(\bm x, \bm z) + \tilde{d}_1(\bm z, \bm x')$. 

This establishes that $\tilde{d}_1$ is a proper distance metric on $M_1$. Now because the composition of injective functions is injective, then each $\tilde{d}_i$ will be a proper distance metric, which proves the result.

\subsection{Combining Predictions}\label{appendix:combining} This section details how to combine the predictions from the Vecchia GPs fitted to different intermediate spaces. We begin with our recommendation, followed by a discussion of the key components. Experimental results supporting this can be found in \cref{appendix:combining_gps_experiment}.

\subsubsection{Recommendation} We recommend combining predictions in $y$-space using the ``Posterior Variance" method in \cref{tab:combining_options}, without the softmax function. This is effective when ensemble members have different noise terms, and no clear method exists for combining them. We also suggest exploring $f$-space combination by finding a principled approach to handling noise terms, especially for extending results to non-Gaussian likelihoods.

\subsubsection{Combining Strategies} When combining predictions from the ensemble of Deep Vecchia GPs, we must decide whether to combine in $y$-space (as in \textcite{liu2020sngp}) or $f$-space (as in \textcite{cohen2020wassersteinBarycenter, deisenroth2015rbcm}), how to compute the unnormalized weights $\beta_k(\bm x^*)$, and whether to use a softmax function for normalizing the weights. If using softmax, a temperature parameter $T$ also needs to be selected.

\subsubsection{Weighting Schemes for GP Ensembles} There are multiple methods for computing $\beta_k$. For example, the softmax-based approach from \textcite{cohen2020wassersteinBarycenter} sets $\beta_k(\bm x^*) \propto \text{exp}(-T \psi_k(\bm x^*))$, where $T$ controls sparsity and $\psi_k(\bm x^*)$ measures the model's confidence. Without softmax, $\beta_k \propto \psi_k(\bm x^*)$, which aligns with product of experts methods. Various choices of $\psi_k(\bm x^*)$ lead to different weighting schemes, such as: $\psi_k(\bm x^*) = 0$ (standard PoE), $\psi_k(\bm x^*) = \sigma_k^2(\bm x^*)$ (higher weight for lower variance), and $\psi_k(\bm x^*) = \frac{1}{2}(\log \sigma^2 - \log \sigma_k^2(\bm x^*))$ (differential entropy weighting from \textcite{cao2014gpoe}). Another option is to use the Wasserstein distance between the prior and posterior \textcite{cohen2020wassersteinBarycenter}.

\section{EXPERIMENT DETAILS}\label{appendix:experiment_details}
This section contains details for the experiments we conducted. This includes details on computing infrastructure, hyperparameter, data splits, baseline implementations and choices specific to each experiment.

\subsection{Environment and Hyperparameters}
All experiments were designed to run efficiently with minimal resources on a machine running Ubuntu 22.04, equipped with an AMD Ryzen 9 5950X, a single NVIDIA 3090 Ti, and 64 GB of RAM. Larger hyperparameter sweeps were conducted on an HTCondor cluster, which consists of various node types and GPUs, providing additional computational power.

Hyperparameters were selected by minimizing either RMSE or jointly minimizing RMSE and NLL on the validation set. For each experiment and model combination, hyperparameter optimization was performed using a budget of 20-200 (depending on the problem/time). For single objective optimization we used a Tree Parzen Estimator as implemented in the Optuna \parencite{akiba2019optuna} plugin for the Python package Hydra \parencite{Yadan2019Hydra}. Likewise, for multiobjective optimization  we used the Optuna implementation of NSGA-II \parencite{nsgaii}. The hyperparameters chosen were those that achieved the highest value of validation RMSE in the single objective tasks. For the multiobjective tasks, we preferred hyperparameters that resulted in better NLL, but never to the extend that RMSE significantly decreased. For any multistage process, we optimized the hyperparameters at each stage independently and set them for any subsequent search.

\subsection{UCI Regression}\label{appendix:uci_details}
Below we provide details on the UCI regression experiment. We begin with the network that is used in \cref{tab:application_uci} and then provide details on the deep GP baseline. 

\subsubsection{Network and Training}
The network used is identical for each UCI regression problem presented in the main text and the weights and activation function are given in \cref{tab:network_params}. For each dataset we chose a batchsize and number of epochs such that the training time was about the same for each task. For three random seeds we then performed train-validation-test splits with 64\%, 16\% and 20\%, respectively. Using the first split we chose around 200 random configurations of the hyperparameters in the bottom of \cref{tab:network_params} to train the network on the training data. We evaluated the MSE on the validation set and chose the the hyperparameters that gave the lowest MSE on the validation set. The weights, splits and hyperparameters were then stored for each of the three random seeds.

\begin{table*}[t]
    \caption{The feed forward neural network from the UCI regression tasks.}
    \label{tab:network_params}
    \centering
    \begin{tabular}{ll}
    \toprule
    \multicolumn{2}{c}{Network Parameters}\\
    \cmidrule(lr){1-2}
    Neurons Per Hidden Layer & 512, 128, 64, 32, 16 \\
    Activation & SELU\\
    &\\
    \multicolumn{2}{c}{Training Hyperparameters}\\
    \cmidrule(lr){1-2}
    Hyperparameters Optimized & learning rate, $\alpha$-drop out, weight decay\\
    
    \bottomrule
    \end{tabular}
\end{table*}

\subsubsection{Deep GP Details}
For the deep GP we used three layer networks. Each hidden layer had dimension $d_x$, where $d_x$ is the dimension of the input for that problem. For example, $d_x = 8$ for kin40k. 

To approximate the posterior of the deep GP we used the method of \textcite{salimbeni2017doubly}, as implemented in GPyTorch \parencite{gardner2018gpytorch}. This inducing point based approach uses mini-batches as well as Monte Carlo to estimate the ELBO. In all instances, we used 100 inducing points per layer and five samples from the variational distribution to esimate the log-likelhood term in the ELBO. We used an RBF automatic relevance determination (ARD) kernel for each GP. 

During training we used a batch size of 1024 for every dataset and set the number of epochs so that training time was between 20-50 minute from start to finish. We used an Adam optimizer with an initial learning rate chosen through a simple grid search for fixed epoch and batch size. The learning rate was decreased during training using cosine annealing with a final learning rate set to be 1e-9. 

The inputs and outputs were scaled to have unit variance and zero mean. During training we observed that the validation RMSE would often plateau for long periods and then suddenly improve. To avoid this we found it was useful to initialize the inducing points by drawing from a standard normal and setting every value whose magnitude was less than one to zero. Additionally, we used a Gamma(9,0.5) prior for all the lengthscales (we have used the shape/rate parameterization of the Gamma distribution). During evaluation, we centered and sclaed the test data using moments calculated on the training data. 

\subsection{S-Curve}\label{appendix:s_curve}
In this experiment we use a feedforward network with three hidden layers that each have two units followed by a SeLU activation function. We utilized the S-curve dataset comprised of 1000 samples \parencite{tenenbaum2000scurve} which can be seen as a 1D curve that is embedded in 3D space. This dataset is illustrated in the top left panel of \cref{fig:s_curve}. Each hidden layer of the network was made up of two units that were followed by the SeLU activation function. The three intermediate representations can be visualized by plotting them directly. We colored the points by the response value in the experiment, and the resulting plot is displayed in the three panels at the bottom of the figure. The top row's remaining two panels show the predictions (and RMSE) from the Deep Vecchia Ensemble and the neural network, respectively. The Deep Vecchia Ensemble utilized all three intermediate representations to form predictions, while the network only used the final intermediate representation. Using the Deep Vecchia Ensemble resulted in an improvement in RMSE.

\subsection{Chemical Property Prediction}\label{appendix:chemical_props}
Below we provide details on data Processing and fine-tuning for the chemical property prediction tasks in \cref{sec:chemical_props}. 
\subsubsection{Data Processing}
In this experiment we use 10 train, validation, and test splits. For each of the 10 train, validation and test splits we followed the same process. To be precise we followed the data processing steps described in Section 2.3 of \textcite{irwin2022chemformer} as closely as possible. Below is a bulleted list of the steps taken to prepare the data for training. The steps are nearly identical to those presented in \textcite{irwin2022chemformer} but with a few extra details provided. 
\begin{itemize}
    \item Convert each SMILES string into its canonical form and find all instances which occur in more than one of the three datasets and place these into the training set. 
    \item Split the remaining observations such that the train, validation and test splits make up 75\%, 10\% and 15\% of the dataset, respectively. 
    \item For each dataset, scale the target value such that the training set has values in zero to one. Scale the validation and test target values by the same amounts. 
    \item Up-sample the ESOL and FreeSolvation datasets by a factor of 2 and 3, respectively. This is only done in the training set used for fine-tuning. 
\end{itemize}

\subsubsection{Fine-Tuning}
For fine-tuning we used the finetuning procedure given in the Chemformer github repo: \href{https://github.com/MolecularAI/Chemformer}{https://github.com/MolecularAI/Chemformer}. We used the default hyper-parameters and chose the model weights such that the validation loss was minimized. The results obtained are similar to those reported in Table 3 of \textcite{irwin2022chemformer}, except for the Free Solvation results. Consistently, for all 10 splits the Free Solvation RMSE we obtained was higher than that reported in the original Chemformer paper.

\subsubsection{Changing SMILES and Forming the Ensemble}\label{appendix:chemical_props_model}
In this problem the SMILES strings are combined with a token that denotes the property being predicted. As an example ``\texttt{C=CC(C)C}" would become ``\texttt{<FreeS>|C=CC(C)C}" for a SMILES string from the FreeSolvation dataset. Then the processed strings are fed through several transformer encoder layers. Now since each SMILES string is potentially a different length, we take the intermediate representation of each encoder layer to be the length 512 vector whose position corresponds to the property being predicted. We now have a network with $L$ layers with well defined intermediate representations and we can proceed to construct a deep Vecchia ensemble as described in \cref{sec:model_construction}.

\section{ADDITIONAL EXPERIMENTAL RESULTS}\label{appendix:more_experiments}
In this section we present additional experimental results which supplement experiments in the main text or present experiments not found in the main text. 

\subsection{UCI Regression}\label{appendix:more_uci_regression_results}
This section contains experimental results that relate to the UCI regression tasks presented in \cref{sec:applications_uci}. This includes more results related to \cref{tab:application_uci}, additional baselines (e.g. NN-ensembles) and other ablation studies.

\subsubsection{Complete UCI Results for SELU Network}
\cref{tab:uci_regression_complete} is the complete version of \cref{tab:application_uci} from the main text, with the protein dataset included. It is worth noting that the results in \cref{tab:uci_regression_complete} are for a network using a SELU activation and the results in \cref{tab:more_uci_regression} are for a network using ReLU activations.

\begin{table}[t]
    \scriptsize
    \caption{\textbf{DVE is consistently competitive with baselines}. NLL $\downarrow$ (left columns) and RMSE $\downarrow$ (right columns) on UCI benchmarks using SELU activation. Upper half represents single pass methods while the lower half represents multiple pass methods.}
    \label{tab:uci_regression_complete}
    \centering
    \begin{tabular}{llll}
    \toprule
    Dataset \\ ($n$,$d$)& Method & NLL $\downarrow$ & RMSE $\downarrow$\\
    \cmidrule(lr){1-1}\cmidrule(lr){2-2}\cmidrule(lr){3-3}\cmidrule(lr){4-4}

            & DVE  & -1.445 (0.011) & 0.057 (0.001)\\
    kin40k  & Neural Net & N/A & 0.069 (0.000) \\
    (40K,8) & Vecchia & -0.028 (0.015)& 0.209 (0.015)\\
            & SVI-GP & -0.093 (0.008) & 0.068 (0.005) \\
            & Deep GP (3) & \textbf{-1.664 (0.027)} & \textbf{0.030 (0.001)}\\
           
    \cmidrule(lr){1-4}

                & DVE         & \textbf{0.350 (0.010)} & 0.345 (0.004)\\
    Elevators   & Neural Net  & N/A & \textbf{0.340 (0.003)} \\
    (16.6K,18)  & Vecchia     & 0.525 (0.015) & 0.414 (0.006)\\
                & SVI-GP      & 0.455 (0.017)& 0.381 (0.007) \\
                & Deep GP (3) & 0.366 (0.029) & 0.346 (0.011)\\
    \cmidrule(lr){1-4}

                & DVE         & -1.021 (0.030) & 0.087 (0.002)\\
    KEGG        & Neural Net  & N/A & \textbf{0.085 (0.002)} \\
    (48.8K,20)  & Vecchia     & -0.987 (0.026) & 0.090 (0.002)\\
                & SVI-GP      & -0.912 (0.033) & 0.098 (0.003) \\
                & Deep GP (3) & \textbf{-1.036 (0.036)} & 0.086 (0.004)\\
    \cmidrule(lr){1-4}

                & DVE         & \textbf{-3.484 (0.054)} & \textbf{0.002 (0.000)}\\
    bike        & Neural Net  & N/A & 0.006 (0.000) \\
    (17.4K, 17) & Vecchia     & -0.239 (0.010) & 0.183 (0.001) \\
                & SVI-GP      & -1.140 (0.047)& 0.073 (0.004) \\
                & Deep GP (3) & -2.806 (0.100) & 0.004 (0.001)\\
    \cmidrule(lr){1-4}

                & DVE         & \textbf{0.646 (0.029)} & \textbf{0.494 (0.014)}\\
    protein     & Neural Net  & N/A & 0.514 (0.042) \\
    (45.7K,9)   & Vecchia     & 0.851 (0.003) & 0.571 (0.002) \\
                & SVI-GP      & 1.083 (0.001) & 0.713 (0.001) \\
                & Deep GP (3) & 0.938 (0.025) & 0.597 (0.018)\\

    \bottomrule
    \end{tabular}
\end{table}

\subsubsection{Additional UCI Baselines}
This experiment aims to answer the following: 
\begin{itemize}
    \item Does DVE work if we change the network?
    \item How does DVE compare with Monte Carlo dropout, deep ensembles, deep kernel learning?
\end{itemize}

To answer these questions we repeat the UCI regression task on a subset of the datasets from \cref{tab:uci_regression_complete} but we replace the SELU activation with a ReLU activation and compare with these alternative baselines. NLL and RMSE results for this experiment are shown in \cref{tab:more_uci_regression}. In the Table ``MC  Dropout (20)" denotes that we use 20 passes through the network for each observation. Likewise, ``Ensemble (5)" denotes an ensemble built using five networks. The subscript for DKL indicates the spectral norm bound used, while the absence of a subscript denotes no spectral normalization. As before the metrics are computed using centered and scaled responses, so the values in \cref{tab:more_uci_regression} and directly comparable to the values in \cref{tab:uci_regression_complete}. 

We can see from \cref{tab:more_uci_regression} that simply adding the DVE on an existing network gives results competitive to ensembles. Stated differently, ensembling layers of a single network can be as effective as ensembling different networks. However, with DVE we don't have to worry about inducing diversity among the models as we simply take the best deterministic network we can find, decompose the layers and fit the intermediate GPs. Furthermore, adding spectral normalization alone to the feature extractor of our DKL variant improves RMSE and NLL compared to the baseline DKL variant. However, DKL-SN does not outperform DVE in terms of NLL on these examples. This outcome is expected, as the network lacks residual connections, preventing the feature extractor from behaving like the constrained feature extractor in a model such as DUE \parencite{amersfoot2021due}. The constrained feature extractor of DUE is crucial for fully leveraging DKL, whereas DVE allows us to bypass this requirement.

\begin{table}[t]
    \scriptsize
    \caption{\textbf{Adding DVE on top of a network consistently achieves competitive performance compared to baselines}. RMSE and NLL for UCI regression tasks using ReLU activation with various baselines for comparison.}
    \label{tab:more_uci_regression}
    \centering
    \begin{tabular}{llll}
    \toprule
    Dataset & Model & NLL $\downarrow$ & RMSE $\downarrow$\\
    \cmidrule(lr){1-1}\cmidrule(lr){2-2}\cmidrule(lr){3-3}\cmidrule(lr){4-4}

            & Neural Net         & N/A& 0.044 (0.001)\\
    kin40k  & DVE                & \textbf{-1.810 (0.007)} &  0.046 (0.001)\\
            & DKL                & -1.511 (0.005) & 0.044 (0.001)\\
            & DKL-SN$_{1.0}$       & -1.400 (0.013) & 0.054 (0.001)\\
            & DKL-SN$_{2.0}$       & -1.544 (0.002) & \textbf{0.042 (0.000)}\\
            & DKL-SN$_{3.0}$     & -1.533 (0.007) & 0.043 (0.001)\\
            & DKL-SN$_{4.0}$       & -1.535 (0.003) & \textbf{0.042 (0.000)}\\
            
    \cmidrule(lr){2-4}
            & Ensemble (5)        & -1.716 (0.026) & \textbf{0.042 (0.002)}\\
            & MC Dropout (20)    & -1.653 (0.019) & 0.052 (0.001)\\
    \cmidrule(lr){1-4}

            & Neural Net         & N/A & 0.014 (0.000)\\
    bike    & DVE                & \textbf{-4.469 (0.060)} & 0.004 (0.001)\\
    
            & DKL                & -2.851 (0.056) & 0.009 (0.001)\\
            & DKL-SN$_{1.0}$       & -1.178 (0.353) & 0.051 (0.004)\\
            & DKL-SN$_{2.0}$       & -3.001 (0.017) & \textbf{0.003 (0.000)}\\
            & DKL-SN$_{3.0}$       & -2.989 (0.017) & 0.005 (0.000)\\
            & DKL-SN$_{4.0}$       & -2.936 (0.033) & 0.007 (0.001)\\
            
                \cmidrule(lr){2-4}
            & Ensemble (5)      & -3.516 (0.234) & 0.010 (0.000) \\
            & MC Dropout (20)    & -2.543 (0.090) & 0.016 (0.001)\\
    \bottomrule
    \end{tabular}
\end{table}

\subsubsection{DVE Outperforms Individual Layers}
This experiment compares the performance of the individual layer-wise models used to construct DVE with DVE itself, which ensembles these layers. We use the same DVE model as in \cref{tab:more_uci_regression}, but now we compare it to the layer-wise Vecchia GPs placed at intermediate layers.

The results are presented in \cref{tab:dve_layerwise_uci_regression}, where Vecchia GP$_{1}$ refers to the Vecchia GP model that utilizes the first intermediate representation (i.e., the first hidden layer). We see that DVE performs favorably compared to the individual layer-wise models across both datasets. Notably, the best-performing individual layer can vary depending on whether RMSE or NLL is considered. In contrast, DVE consistently delivers strong performance for both metrics while sidestepping the challenge of selecting the optimal layer.

\begin{table}[t]
    \scriptsize
    \caption{\textbf{DVE's performance matches or exceeds that of the best individual layer-wise model in the ensemble, sidestepping the need to identify the optimal layer.}. RMSE and NLL for UCI regression tasks using ReLU activation, comparing DVE to the individual layer-wise models that are combined to form DVE.}
    \label{tab:dve_layerwise_uci_regression}
    \centering
    \begin{tabular}{llll}
    \toprule
    Dataset & Model & NLL $\downarrow$ & RMSE $\downarrow$\\
    \cmidrule(lr){1-1}\cmidrule(lr){2-2}\cmidrule(lr){3-3}\cmidrule(lr){4-4}

    kin40k  & DVE                & \textbf{-1.810 (0.007)} &  \textbf{0.046 (0.001)}\\
            & Vecchia GP$_{1}$   & -0.857 (0.085) &  0.090 (0.004)\\
            & Vecchia GP$_{2}$   & -1.575 (0.039) &  0.053 (0.001)\\
            & Vecchia GP$_{3}$   & -1.723 (0.022) &  0.048 (0.002)\\
            & Vecchia GP$_{4}$   & -1.726 (0.068) &  0.047 (0.003)\\
    \cmidrule(lr){1-4}

    bike    & DVE                & -4.469 (0.060) &  0.004 (0.001)\\
            & Vecchia GP$_{1}$   & -4.361 (0.153) &  \textbf{0.003 (0.001)}\\
            & Vecchia GP$_{2}$   & -4.423 (0.083) &  0.005 (0.002)\\
            & Vecchia GP$_{3}$   & \textbf{-4.504 (0.049)} &  0.004 (0.002)\\
            & Vecchia GP$_{4}$   & -4.454 (0.117) &  0.003 (0.001)\\
            
    \bottomrule
    \end{tabular}
\end{table}

\subsubsection{Intermediate Layers Preserve Information}
This experiment examines how intermediate representations preserve information relevant to estimating epistemic uncertainty, even when that information is lost in the final layer. We build on the binary classification task from \textcite{amersfoot2020DUQ}, where two-dimensional points are classified based on the sign of their first coordinate. A network trained on this task should ignore perturbations in the second coordinate, treating out-of-distribution points similarly to training data. However, in active learning, we often want to identify and label such points to reduce epistemic uncertainty, creating a tension between classification accuracy and sensitivity to novel points.  

We reuse the implementation from \textcite{jimenez2025psc}, including their data generation and plotting code, with slight modifications to the network architecture. The network consists of layers with dimensions [8, 8, 4, 4, 4, 4], residual connections, and Leaky-ReLU activations throughout. The dataset contains $N = 2000$ samples of two-dimensional points ($d = 2$) drawn from a Gaussian distribution with variance $\sigma = 0.3$. The labels are determined by the sign of the first coordinate ($y = \text{sign}(x_1)$), with label noise introduced by flipping labels with probability $\epsilon_p = 0.005$. The test set is generated similarly but without noise.  

We train two networks, one with spectral normalization and one without, and examine their intermediate representations by projecting them onto the first two principal components. We visualize the projections using the full test set and two perturbations of a test point: one class-relevant (blue) and one irrelevant (red), which perturbs the second coordinate.  

\begin{figure}
    \centering
    \includegraphics[width=1.0\linewidth]{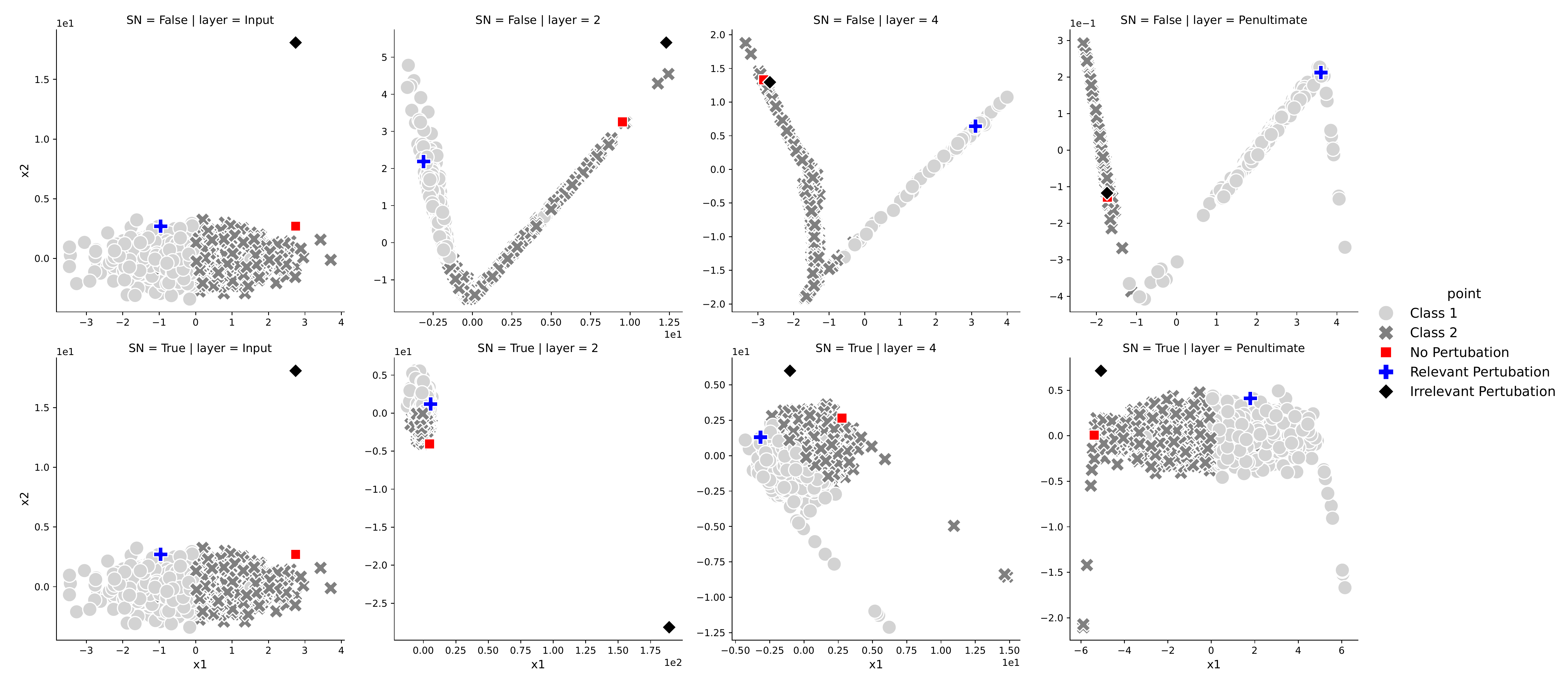}
    \caption{\textbf{Without spectral normalization, intermediate layers are sensitive to perturbations irrelevant to the class label}, as shown in the second panel from the right in the top row. This sensitivity persists even though later layers in the same network collapse the representation of the irrelevant perturbation to resemble in-distribution data. In contrast, with spectral normalization, the perturbation is preserved throughout the network (bottom row). However, DVE can leverage the earlier layers, enabling us to work with networks that are not trained with spectral normalization.}
    \label{fig:collapse}
\end{figure}

The results, shown in \cref{fig:collapse}, demonstrate that without spectral normalization, the irrelevant perturbation collapses to resemble in-distribution points, whereas with spectral normalization, the point remains distinguishable from in-distribution data even in the penultimate layer. Early layers in the network without spectral normalization (e.g., layer 1) preserve information about the perturbation, suggesting they help estimate epistemic uncertainty.  

A similar observation was made in \textcite{jimenez2025psc}, which showed that intermediate representations, especially from the penultimate layer of networks with spectral normalization, can distinguish in- and out-of-distribution data.

\subsection{Combining Layer-Wise Predictions}\label{appendix:combining_gps_experiment}
In this section, we empirically investigate how to optimally combine the predictions from the layer-wise GPs in the DVE. As discussed in \cref{appendix:combining}, there are several design choices involved in combining the predictions from the layer-wise Vecchia GPs, and this study serves to guide us toward our final recommendation.

To assess the impact of each design choice, we measure the DVE's validation RMSE and NLL on two datasets from the UCI regression tasks: bike and kin40k. In each setting, we begin by fitting the layer-wise GPs to the same training data that was used to train the network. We then use the validation set to generate GP predictions for each layer and then combine them using all meaningful combinations of the options listed in \cref{tab:combining_options}. For example, we combine Wasserstein weighting in $f$-space using softmax, with the temperature set to 3. For each combination, then form the combined prediction and measure performance. Training is performed on the training set, and results are evaluated on the validation set. 

The RMSE and NLL values for both bike and kin40k when combining without softmax are shown in \cref{fig:f_vs_y_no_softmax}. The RMSE and NLL values when using softmax for bike and kin40k are presented in \cref{fig:f_vs_y_softmax_bike} and \cref{fig:f_vs_y_softmax_kin40k}, respectively. Below, we analyze the results in relation to various concepts and discuss how our observations compare to existing literature on combining model predictions.

\subsubsection{Combining in $f$-Space or $y$-Space}\label{appendix:f_vs_y}
From \cref{fig:f_vs_y_no_softmax}, \cref{fig:f_vs_y_softmax_bike} and \cref{fig:f_vs_y_softmax_kin40k} we see that the NLL is lower when combining in $y$-space. This holds for every combination of combining  method, combining space and temperature value. While the story is not as clear when considering RMSE, the results are comparable when combining in $y$-space or $f$-space. Therefore, we suggest combining in $y$-space when using a simple average of the noise terms from each Vecchia GP. However, these results will certainly change if the noise terms from each Vecchia GP are combined in a more principled way, and especially if the noise terms influence the weight being assigned to each GP in the ensemble. A complete investigation of combining in $f$-space for the deep Vecchia ensemble would likely result in further improvement over what is presented here. 

These results are unique because, for PoE models relying on disjoint partitions of the data, it has been shown that combining in $f$-space can be preferable. Additionally, for non-Gaussian likelihoods it necessary to combine in $f$-space as combining in $y$-space may result in intractable inference \parencite{cohen2020wassersteinBarycenter}. However, the effect of combining in $y$-space vs.\ $f$-space is different here than what was observed in \textcite{cohen2020wassersteinBarycenter}, as we do not use GPs on disjoint partitions of the data. Additionally, each Vecchia GP likelihood has its own value for the observation noise. For simplicity we take an average of these observation noise terms, but a more principled approach would likely improve the results of combining in $f$-space. We do not address this issue here and simply compare combining in $y$-space to combining in $f$-space with a simple average of observation noise terms. 

\subsubsection{Softmax or No Softmax}
\begin{figure*}
    \centering
    \includegraphics[width =0.9\linewidth]{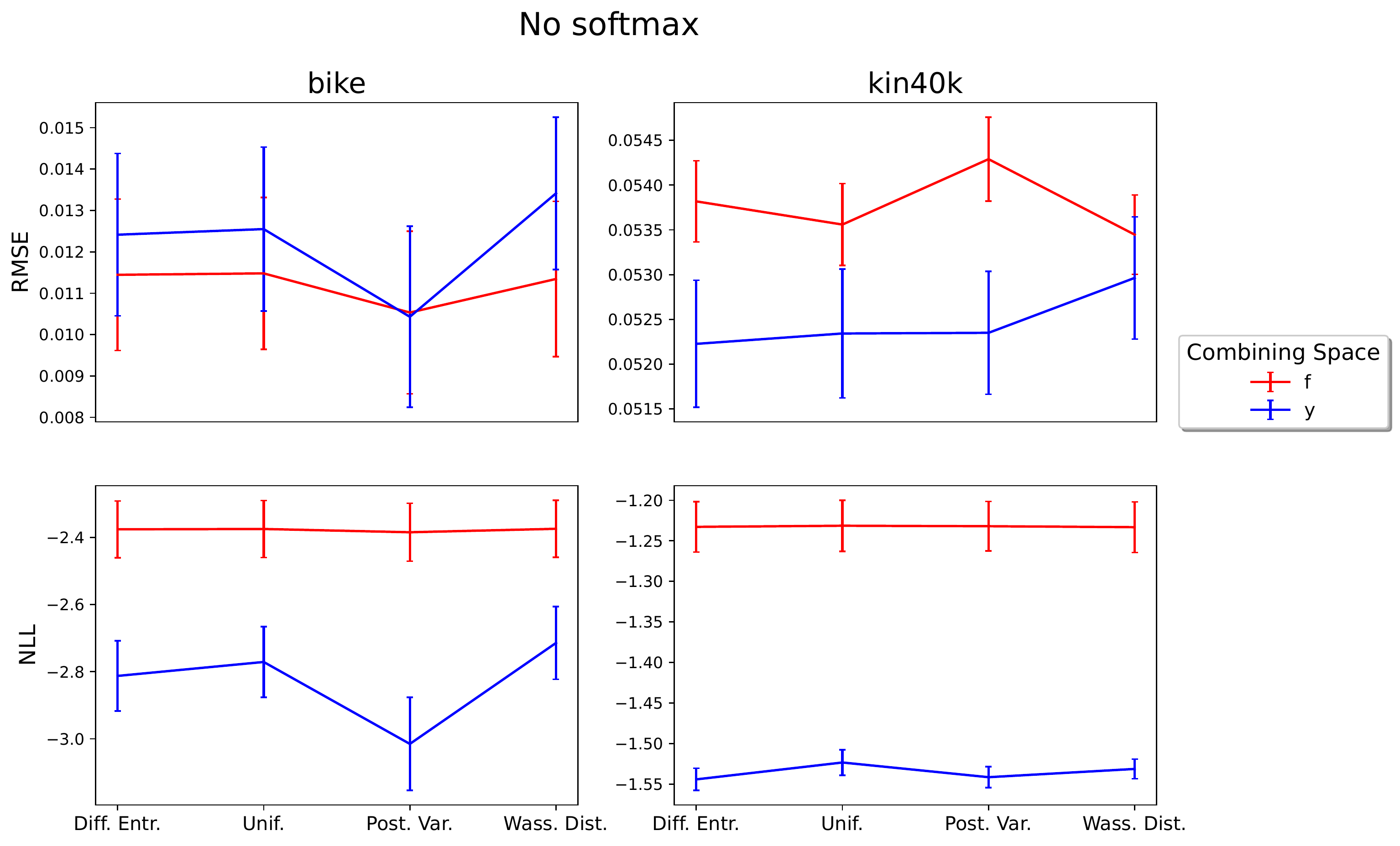}
    \caption{The RMSE and NLL (lower is better) for Deep Vecchia ensemble using different combining methods in both $f$-space (red) and $y$-space (blue) without using the temperature endowed softmax. The left column shows the results on bike and the right columns show the results on kin40k. The vertical bars on each plot repersent one standard deviation (from three repeats). In terms of NLL, combing in $y$-space has better results for every combining method. The results for RMSE are not as clear.}
    \label{fig:f_vs_y_no_softmax}
\end{figure*}

\begin{figure*}
    \centering
    \includegraphics[width =0.9\linewidth]{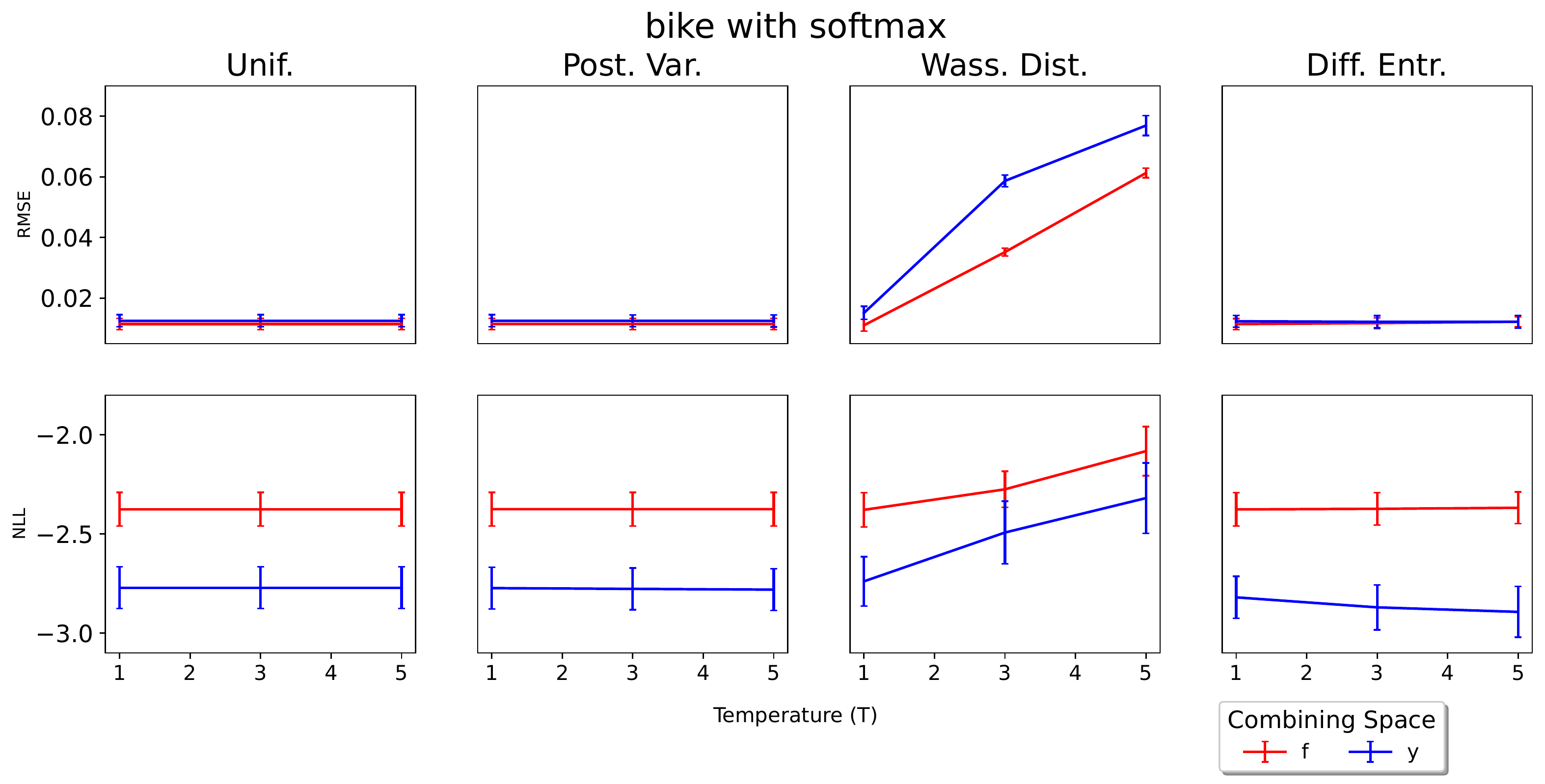}
    \caption{The RMSE and NLL vs temperature (T) for Deep Vecchia ensemble using different combining methods in both $f$-space (red) and $y$-space (blue). The results shown are all for the \textbf{bike} dataset. The vertical bars on each plot represent one standard deviation (from three repeats). It appears that increasing the value of $T$ had meaningful change on the NLL of Wasserstein distance and differential entropy, but not on posterior variance or uniform weighting. For RMSE only Wasserstein distance changed as we varied $T$. }
    \label{fig:f_vs_y_softmax_kin40k}
\end{figure*}

\begin{figure*}
    \centering
    \includegraphics[width =.9\linewidth]{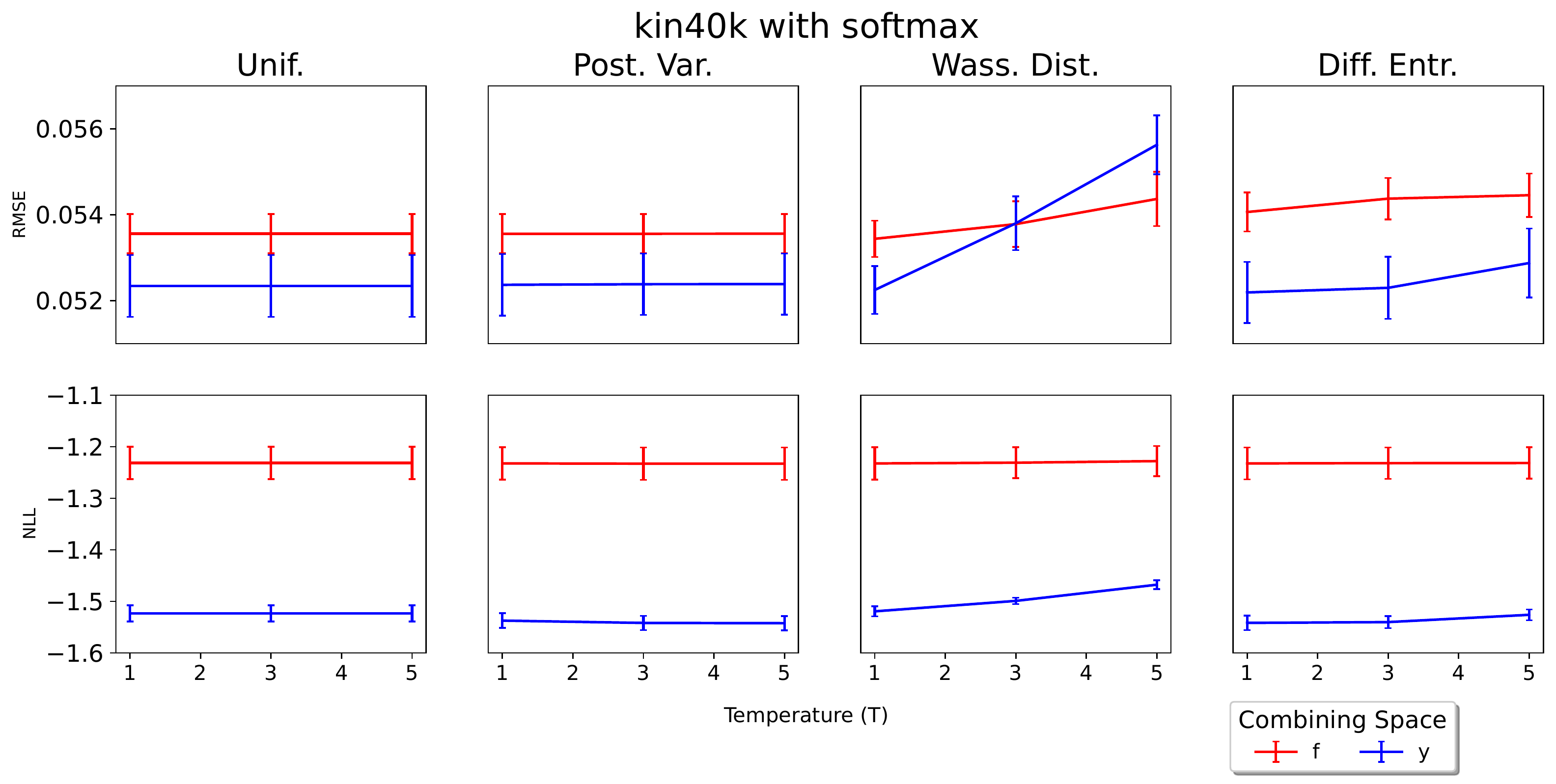}
    \caption{The RMSE and NLL vs temperature (T) for Deep Vecchia ensemble using different combining methods in both $f$-space (red) and $y$-space (blue). The results shown are all for the kin40k dataset. The vertical bars on each plot represent one standard deviation (from three repeats). Just as in \cref{fig:f_vs_y_softmax_bike} NLL changed for Wasserstein distance and differential entropy as we varied $T$, but not for other methods. RMSE for Wasserstein distance and differential entropy were the only two that changed as we increased $T$. }
    \label{fig:f_vs_y_softmax_bike}
\end{figure*}

For the bike dataset, the NLL and RMSE values for ``Posterior Variance" in \cref{fig:f_vs_y_no_softmax} are as good as any any NLL and RMSE value given in \cref{fig:f_vs_y_softmax_bike} (considering the error bars of each plot). This implies there is no value of $T$ for the softmax that outperforms no softmax on the bike dataset. The same observation holds when comparing kin40k results without softmax (\cref{fig:f_vs_y_no_softmax}) to those using softmax (\cref{fig:f_vs_y_softmax_kin40k}). Furthermore, without the use of a softmax function we do not need to choose the value of the hyperparameter $T$. Therefore, for our work we did not use the softmax function. We believe that similar results may may hold for other datasets when using the deep Vecchia ensemble and combining in $f$-space as we have. Again these results may change if the Veccchia GP noise terms are combined differently, but we do not investigate that here.

\begin{table*}[t]
    \caption{Different options for combining Deep Vecchia Ensemble predictions. Weigthing formulas correspond to unnormalized weights.}
    \label{tab:combining_options}
    \centering
    \begin{tabular}{ll}
    \toprule
    Component & Options\\
    \cmidrule(lr){1-2}
    Weighting  & Uniform $\beta_k(\bm x_*) = 1$\\
    & Posterior Variance $\beta_k(\bm x_*) = 1/\sigma_j^2(\bm x_*)$\\
    & Differential Entropy $\beta_k(\bm x_*) = 0.5 (\sigma_{**}^2- \sigma_j^2(\bm x_*))$\\
    & Wasserstein Dist. $\mu_j(\bm x_*)^2 + (\sigma_{**}^2- \sigma_j^2(\bm x_*))^2$\\
    \cmidrule(lr){1-2}
    Combining space & $f$-space / $y$-space \\
    \cmidrule(lr){1-2}
    softmax & True / False \\
    \cmidrule(lr){1-2}
    Temperature (T) & 1, 3, 5\\
    \bottomrule
    \end{tabular}
\end{table*}

